\documentclass[10pt, compsoc, conference, a4paper]{IEEEtran}

\usepackage{ifthen}
\usepackage[normalem]{ulem} % for \sout
\usepackage{xcolor}
\usepackage{amssymb}

\newboolean{showedits}
\setboolean{showedits}{true} % toggle to show or hide edits
\ifthenelse{\boolean{showedits}}
{
	\newcommand{\del}[1]{\textcolor{red}{\sout{#1}}} % please delete
}{
	\newcommand{\del}[1]{} % please delete
	
}

\newboolean{showcomments}
\setboolean{showcomments}{false} % ATTN: THIS TURNS ON THE ANNOTATIONS
\newcommand{\id}[1]{$-$Id: scgPaper.tex 32478 2010-04-29 09:11:32Z oscar $-$}

\ifthenelse{\boolean{showcomments}}
{\newcommand{\nbc}[3]{
 {\colorbox{#3}{\bfseries\sffamily\scriptsize\textcolor{white}{#1}}}
 {\textcolor{#3}{\sf\small$\blacktriangleright$\textit{#2}$\blacktriangleleft$}}}
 }
{\newcommand{\nbc}[3]{}
 \renewcommand{\del}[1]{} % please delete
 }

\definecolor{ibcolor}{rgb}{0.9,0.5,0}
\definecolor{cfcolor}{rgb}{0,0.5,0.9}
\definecolor{tdcolor}{rgb}{1.0,0,0}

\usepackage{amsmath}
\usepackage{graphicx}
\usepackage{enumitem}
\usepackage{balance}
\usepackage{array}

\usepackage{float}
\usepackage{hyperref}
\usepackage{dirtytalk}

\usepackage[linesnumbered]{algorithm2e}

\SetCommentSty{mycommfont}

\usepackage[page]{appendix}

\begin{document}
\title{Mitigating Sybils in Federated Learning Poisoning}

\author{\IEEEauthorblockN{Clement Fung}
\IEEEauthorblockA{University of British Columbia\\
clement.fung@alumni.ubc.ca}
\and
\IEEEauthorblockN{Chris J.M. Yoon}
\IEEEauthorblockA{University of British Columbia\\
yoon@alumni.ubc.ca}
\and
\IEEEauthorblockN{Ivan Beschastnikh}
\IEEEauthorblockA{University of British Columbia\\
bestchai@cs.ubc.ca}}

\maketitle

\begin{abstract}
Machine learning (ML) over distributed multi-party data is required
for a variety of domains. Existing approaches, such as federated
learning, collect the outputs computed by a group of devices at a
central aggregator and run iterative algorithms to train a globally
shared model. Unfortunately, such approaches are susceptible to a
variety of attacks, including model poisoning, which is made
substantially worse in the presence of sybils.

In this paper we first evaluate the vulnerability of federated
learning to sybil-based poisoning attacks. We then describe
\emph{FoolsGold}, a novel defense to this problem that identifies
poisoning sybils based on the diversity of client updates in the
distributed learning process. Unlike prior work, our system does not
bound the expected number of attackers, requires no auxiliary
information outside of the learning process, and makes fewer
assumptions about clients and their data.

In our evaluation we show that FoolsGold exceeds the capabilities of
existing state of the art approaches to countering sybil-based
label-flipping and backdoor poisoning attacks. Our results hold for
different distributions of client data, varying poisoning targets, and
various sybil strategies. 
\end{abstract}

\maketitle

%!TEX root = paper.tex
% ^enables bibliography plugin
%%%%%%%%%%%%%%%%%%%%%%%%%%%%%%%%%%%%%%%%%%%%%%%%%%%%%%%%%%%%%%%%%%
\section{Introduction}
\label{sec:intro}
%%%%%%%%%%%%%%%%%%%%%%%%%%%%%%%%%%%%%%%%%%%%%%%%%%%%%%%%%%%%%%%%%%

To train multi-party machine learning (ML) models from user-generated
data, users must provide and share their training data, which can be 
expensive or privacy-violating. Federated learning~\cite{McMahan:2017,
Gboard:2017} is a recent solution to both problems: while training
across mobile devices, data is kept on the client device and only
model parameters are transferred to a central aggregator. This allows
clients to compute their model updates locally and independently,
while maintaining a basic level of privacy.

% Federated learning can be
% further augmented with differential privacy~\cite{Geyer:2017} and
% secure aggregation~\cite{Bonawitz:2017} to provide additional
% client-side privacy and security.

However, federated learning introduces a risky design tradeoff:
clients, who previously acted only as passive data providers, can now
observe intermediate model state and contribute arbitrary updates as
part of the decentralized training process. This creates an
opportunity for malicious clients to manipulate the training process
with little restriction. In particular, adversaries posing as honest
clients can send erroneous updates that maliciously influence the
properties of the trained model, a process that is known as 
\emph{model poisoning}.

Poisoning attacks have been well explored in centralized settings and
two prevalent examples are \emph{label-flipping 
  attacks}~\cite{Biggio:2012, Huang:2011} and \emph{backdoor 
  attacks}~\cite{Bagdasaryan:2018, Chen:2017, Gu:2017}. In both types
of attacks, the larger the proportion of poisoning data, the higher
the attack effectiveness. In a federated learning context, where each
client maintains data locally, an adversary can naturally increase
their attack effectiveness by using \emph{sybils}~\cite{Douceur:2002}.

%% While the concept of sharing poisoned data
%% %
%% \ivan{'sharing poisoned data'? not mentioned above and I'm not sure
%%   what sharing means here.}
%% %
%% does not apply to federated
%% learning, the same strategy can be applied through
%% sybils~\cite{Douceur:2002}. 

%% Sybils have been a common strategy for adversaries to increase their
%% effectiveness in crowd-sourced systems. For example, using a mobile
%% device simulator, an adversary can simulate fake accounts to
%% maliciously manipulate the result of a crowd-sourced
%% computation~\cite{Wang:2016}. To perform a similar attack on federated
%% learning, an adversary can equip multiple user accounts with poisoned
%% data, gain admission to the system, and contribute poisoned updates to
%% the federated learning model.

%%%%%%%%%%%%%%%%%%%%%%%%%%%%%%%%%%%%%%%%%%%%%%%%%%%%%%%%%%%%
\begin{table}[t]
\caption{The accuracy and attack success rates for baseline (no
  attack), and attacks with 1 and 2 sybils in a federated learning
  context with MNIST dataset.}
\label{tab:fedattack}
\centering
\begin{tabular}{r|ccc}
  & Baseline & Attack 1 & Attack 2 \\
 \hline
 \# honest clients & 10 & 10 & 10 \\
 \hline
 \# malicious sybils & 0 & 1 & 2 \\
 \hline
 Accuracy (digits: 0, 2-9) & 90.2\% & 89.4\% &
 88.8\% \\
 \hline
 Accuracy (digit: 1) & 96.5\% & 60.7\% &
 0.0\% \\
 \hline
 \textbf{Attack success rate} & 0.0\% & 35.9\% &
 96.2\% \\
 \hline
\end{tabular} 
\end{table}
%%%%%%%%%%%%%%%%%%%%%%%%%%%%%%%%%%%%%%%%%%%%%%%%%%%%%%%%%%%%

\begin{figure*}[t]
    \includegraphics[width=.9\linewidth]{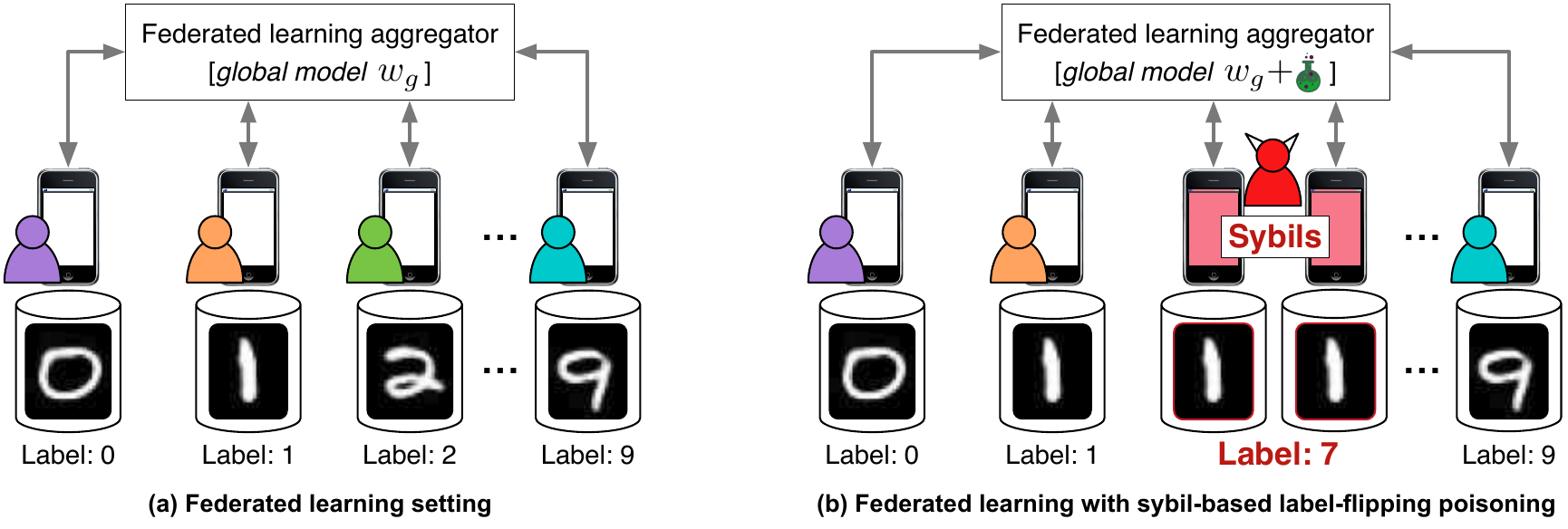}
    \caption{Federated learning with and without colluding sybils
      mounting a sybil-based poisoning attack. In the attack (b) two
      sybils poison the model by computing over images of 1s with the
      (incorrect) class label 7.
    }
    \label{fig:fedattack}
\end{figure*}

\vspace{.2em}
\noindent \textbf{A federated learning poisoning experiment.} We 
illustrate the vulnerability of federated learning to sybil-based
poisoning with three experiments based on the setup in 
Figure~\ref{fig:fedattack} and show the results in  
Table~\ref{tab:fedattack}. First, we recreate the baseline evaluation
in the original federated learning paper~\cite{McMahan:2017} and
train an MNIST~\cite{Lecun:1998} digit classifier across non-IID
data sources (Figure~\ref{fig:fedattack}(a) and Baseline column in 
Table~\ref{tab:fedattack}). We train a softmax classifier across ten
honest clients, each holding a single digit partition of the original
ten-digit MNIST dataset. Each model is trained for 3000 synchronous
iterations, in which each client performs a local SGD update using a
batch of 50 randomly sampled training examples.

%%%%%%%%%%%%%%%%%%%%%%%%%%%%%%%%%%%%%%%%%%%%%%%%%%%%%%%%
%%%%%%%%%%%%%%%%%%%%%%%%%%%%%%%%%%%%%%%%%%%%%%%%%%%%%%%%

We then re-run the baseline evaluation with a label-flipping 1-7
poisoning attack~\cite{Biggio:2011}: each malicious client in the
system has a poisoned dataset of 1s labeled as 7s 
(Figure~\ref{fig:fedattack}(b)). A successful attack generates a model
that is unable to correctly classify 1s and incorrectly predicts them
to be 7s. We define the \emph{attack success rate} as the proportion
of 1s predicted to be 7s by the final model in the test set. We
perform two experiments, in which 1 or 2 malicious sybil clients
attack the shared model (Attack 1 and Attack 2 in 
Table~\ref{tab:fedattack}).

Table~\ref{tab:fedattack} shows that with only 2 sybils, 96.2\% of
1s are predicted as 7s in the final shared model. Since one honest
client held the data for digit 1, and two malicious sybils held the
poisoned 1-7 data, the malicious sybils were twice as influential on
the shared model as the honest client. This attack illustrates a
problem with federated learning: \emph{all clients have equal
influence} on the system\footnote{The only weight in federated
learning is the number of examples that each client possesses, but
this weight is easy for adversaries to inflate by generating more data
or by cloning their dataset.}. %% The number of sybils required by the
%% adversary for successful poisoning is unknown, but
As a result, regardless of the number of honest clients in a federated
learning system, with enough sybils an adversary may overpower these
honest clients and succeed in poisoning the model.

% generating sufficiently many sybils.
% gain enough influence to 

Known defenses to poisoning attacks in a centralized context, such as
robust losses~\cite{Han:2016} and anomaly
detection~\cite{Rubinstein:2009}, assume control of the clients or
explicit observation of the training data. Neither of these
assumptions apply to federated learning in which the server only
observes model parameters sent as part of the iterative ML algorithm.
Therefore, \emph{these defenses cannot be used in federated learning}.

% One may consider that a simple outlier detection method is sufficient
% to detect and remove sybils, but this is difficult for a variety of
% reasons. Since the server is unable to view the training data of
% clients in federated learning and does not possess a validation
% dataset, they cannot easily verify which incoming parameter updates
% are genuine. Furthermore, these updates are a product of a random
% stochastic process and will exhibit high variance. The key insight in
% this work is that when a shared model is being manipulated by a group
% of sybils, they will, in expectation over the entire training process,
% contribute updates towards a shared specific malicious objective,
% exhibiting behavior that is more similar amongst themselves than what
% is naturally observed in the multi-party ML setting.

Poisoning defenses that do apply to federated learning require
explicit bounds on the expected number of sybils~\cite{Blanchard:2017, 
Shen:2016}. To the best of our knowledge, we are the first to mitigate
sybil-based poisoning attacks without an explicit parameter for the
number of attackers.

Since the federated learning server is unable to view client training
data and does not have a validation dataset, the server cannot easily
verify which client parameter updates are genuine. Furthermore, these
updates are a product of a random stochastic process and will exhibit
high variance. We propose \textbf{FoolsGold}: a new defense against
federated learning sybil attacks that adapts the learning rate of
clients based on \emph{contribution similarity}. The insight in this
work is that when a shared model is manipulated by a group of sybils,
they will, in expectation over the entire training process, contribute
updates towards a specific malicious objective, exhibiting behavior
that is more similar than expected. 

FoolsGold defends federated learning from attacks performed by an
arbitrary number of sybils with minimal change to the original
federated learning procedure. Moreover, FoolsGold does not assume a
specific number of attackers. We evaluate FoolsGold on 4 diverse data
sets (MNIST~\cite{Lecun:1998}, VGGFace2~\cite{Cao:2018}, 
KDDCup99~\cite{Dua:2017}, Amazon Reviews~\cite{Dua:2017}) and 3 model types
(1-layer Softmax, Squeezenet, VGGNet) and find that our approach mitigates
label-flipping and backdoor attacks under a variety of conditions,
including different distributions of client data, varying poisoning
targets, and various sybil strategies.

\noindent In summary, we make the following contributions:
\begin{itemize}[label=$\star$]

    \item We consider sybil attacks on federated learning
      architectures and show that existing defenses against malicious
      adversaries in ML (Multi-Krum~\cite{Blanchard:2017} and 
      RONI~\cite{Barreno:2010} (shown in Appendix~B)) 
      are inadequate.

% \ref{sec:extraeval}
   
    \item We design, implement, and evaluate a novel defense against
      sybil-based poisoning attacks for the federated learning setting
      that uses an adaptive learning rate per client based on
      inter-client contribution similarity.

    \item In this context, we discuss optimal and intelligent attacks
    that adversaries can perform, while suggesting possible directions
    for further mitigation.

\end{itemize}

%!TEX root = paper.tex
%%%%%%%%%%%%%%%%%%%%%%%%%%%%%%%%%%%%%%%%%%%%%%%%%%%%%%%%%%%%%%%%%%
\section{Background}
\label{sec:background}
%%%%%%%%%%%%%%%%%%%%%%%%%%%%%%%%%%%%%%%%%%%%%%%%%%%%%%%%%%%%%%%%%%

\noindent \textbf{Machine Learning (ML).} 
Many ML problems can be represented as the minimization of a loss
function in a large Euclidean space. For an ML binary classification
task that predicts a discrete binary output; more prediction errors
result in a higher loss. Given a set of training data and a proposed
model, ML algorithms \emph{train}, or find an optimal set of
parameters, for the given training set. 

\noindent \textbf{Stochastic gradient descent (SGD).} 
One approach in ML is to use stochastic gradient descent
(SGD)~\cite{Bottou:2010}, an iterative algorithm that selects a batch
of training examples, uses them to compute gradients on the parameters
of the current model, and takes gradient steps in the direction that
minimizes the loss function. The algorithm then updates the model
parameters and another iteration is performed. SGD is a general
learning algorithm that can be used to train a
variety of models, including deep learning models~\cite{Bottou:2010}. We
assume SGD as the optimization algorithm in this paper.

In SGD, the model parameters $w$ are updated at each iteration $t$ as
follows:
\begin{align*}
  w_{t+1} = w_t - \eta_t(\lambda w_t + \frac{1}{b} \sum_{(x_i, y_i)\in
  B_t} \nabla l(w_t, x_i, y_i)) &&&& (1)
\end{align*}
where $\eta_t$ represents a local learning rate, $\lambda$ is a
regularization parameter that prevents over-fitting, $B_t$ represents
a gradient batch of training data examples $(x_i, y_i)$ of size $b$
and $\nabla l$ represents the gradient of the loss function. 

Batch size is what separates SGD from its counterpart, gradient descent
(GD). In GD, the entire dataset is used to compute the gradient
direction, but in SGD, a subset is used at each iteration. This subset
may be selected in a pre-determined order or sampled randomly, but the
overall effect of SGD is that the gradient directions seen over time
vary (and have higher variance as the batch size $b$ decreases). In
practice, SGD is preferred to GD for several reasons: it is less
expensive for large datasets and theoretically scales to datasets of
infinite size. 

A typical heuristic involves running SGD for a fixed number of
iterations or halting when the magnitude of the gradient falls below a
threshold. When this occurs, model training is completed and
the parameters $w_t$ are returned as the optimal model $w^*$.

\noindent \textbf{Federated learning~\cite{McMahan:2017}.} 
In FoolsGold, we assume a standard federated learning context, in
which data is distributed across multiple data owners and cannot be
shared. The distributed learning process is performed in
\textit{synchronous update rounds} over a set of clients, in which a
weighted average of the $k$ client updates, based on their
proportional training set size $n_k$ out of $n$ total examples, is
applied to the model atomically.
\begin{align*}
  w_{g,t+1} &= w_{g,t} + \sum_k \frac{n_k}{n} \Delta_{k,t}
\end{align*}
Even if the training data is distributed such that it is not
independent and identically distributed (non-IID), federated learning
can still attain convergence. For example, federated learning can
train an MNIST~\cite{Lecun:1998} digit recognition classifier in a
setting where each client only held data for 1 of the digit classes 
(0-9), as in Figure~\ref{fig:fedattack}(a).

Federated learning comes in two forms: FEDSGD, in which each client
sends every SGD update to the server, and FEDAVG, in which clients
locally batch multiple iterations of SGD before sending updates to the
server, which is more communication efficient. We show that FoolsGold
can be applied successfully on both algorithms.

Federated learning also enables model training across a set of clients
with highly private data. For example, differentially 
private~\cite{Geyer:2017} and securely aggregated~\cite{Bonawitz:2017}
additions have been released, yet the use of federated learning in
multi-party settings still presents new privacy and security
vulnerabilities~\cite{Bagdasaryan:2018, Hitaj:2017, Nitin:2018}. 
In this paper we design FoolsGold to address sybil-based poisoning attacks.

\noindent \textbf{Targeted poisoning attacks on ML.} 
In a targeted poisoning attack~\cite{Biggio:2011,
Mozaffari-Kermani:2015}, an adversary meticulously creates poisoned
training examples and inserts them into the training data set of an
attacked model. This is done to increase/decrease the probability of
the trained model predicting a targeted example as a targeted 
class~\cite{Huang:2011} (see Figure~\ref{fig:fedattack}(b)). For
example, such an attack could be mounted to avoid fraud detection or
to evade email spam filtering~\cite{Nelson:2008}. 

In FoolsGold, we consider targeted attacks of two types. In
\emph{label-flipping attacks}~\cite{Biggio:2012}, the labels of honest training
examples of one class are flipped to another class while the features
of the data are kept unchanged. In \emph{backdoor 
attacks}~\cite{Bagdasaryan:2018, Gu:2017}, single features or small
regions of the original training data are augmented with a secret
pattern and relabeled. The pattern acts as a trigger for the target
class, which is exploited by an attacker. Generally, it is important for a
poisoning attack to not significantly change the prediction outcomes of
other classes. Otherwise, the attack will be detected by users of the
shared model.

\begin{figure}[t]
    \includegraphics[width=.9\linewidth]{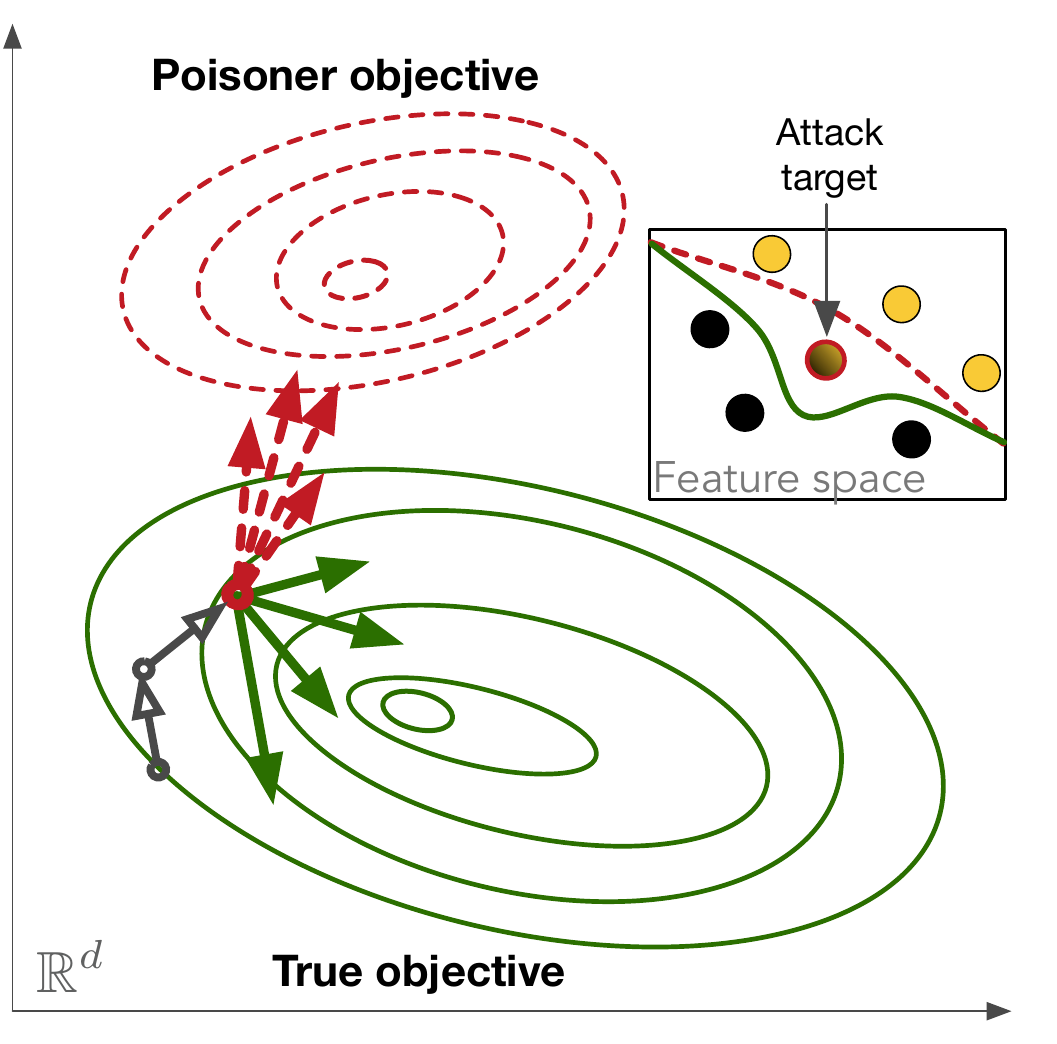}
    \caption{Targeted poisoning attack in SGD. The dotted red vectors
      are sybil contributions that drive the model towards a poisoner
      objective. The solid green vectors are contributed by honest
      clients that drive towards the true objective.}
    \label{fig:sgd-detail}
\end{figure}

In federated learning, the aggregator does not see any training data,
and we therefore view poisoning attacks from the perspective of
model updates: a subset of updates sent to the model at any given
iteration are poisoned~\cite{Blanchard:2017}. This is functionally
identical to a centralized poisoning attack in which a subset of the
total training data is poisoned. Figure~\ref{fig:sgd-detail}
illustrates a targeted poisoning attack in a federated learning
context.

\noindent \textbf{Sybil attacks.} 
A system that allows clients to join and leave is susceptible to 
sybil attacks~\cite{Douceur:2002}, in which an adversary gains
influence by joining a system under multiple colluding aliases. In
FoolsGold, we assume that adversaries leverage sybils to mount more
powerful poisoning attacks on federated learning.

%%%%%%%%%%%%%%%%%%%%%%%%%%%%%%%%%%%%%%%%%%%%%%%%%%%%%%%%

%!TEX root = paper.tex
%%%%%%%%%%%%%%%%%%%%%%%%%%%%%%%%%%%%%%%%%%%%%%%%%%%%%%%%%%%%%%%%%%
%\section{Threat context}
%\label{sec:threat}
%%%%%%%%%%%%%%%%%%%%%%%%%%%%%%%%%%%%%%%%%%%%%%%%%%%%%%%%%%%%%%%%%%

\section{Assumptions and threat model}
\label{sec:threat}

\noindent \textbf{Setting assumptions.} 
We are focused on federated learning and therefore assume that data is
distributed and hidden across clients. The adversary can only access and
influence ML training state through the federated learning API, and
adversaries cannot observe the training data of other honest clients. 

This means that by observing the total change in model state,
adversaries can observe the total averaged gradient across all
clients, but they cannot view the gradient of any individual honest
client. 

On the server side of the algorithm, we assume that the aggregator is
uncompromised and is not malicious. Similarly, we assume that some number
of honest clients, who possess training data which opposes the attacker's
poisoning goal, participate in the system. More
precisely, our solution requires that every class defined by the model
is represented in the data of at least one honest client. Without
these honest clients, no contribution-based defense is possible since
the model would be unable to learn anything about these classes in the
first place.

Secure-aggregation for federated learning provides better
privacy by obfuscating individual client updates~\cite{Bonawitz:2017}.
We assume that these types of obfuscations are not used, and that the
central server is able to observe any individual client's model update
at each iteration. 

% This is more of a RW paragraph. And we already discuss this in RW/cutting.
%
% \clement{The paragraph below is moved from intro}
%% Additionally, alternative solutions to sybil attacks involve detecting
%% malicious users and removing their influence. This leverages social
%% network information~ \cite{Tran:2009} or incorporates auxiliary
%% information about users into the algorithm~\cite{Viswanath:2015,
%% Wang:2016}. In our work, we do not make such assumptions.

\noindent \textbf{Poisoning attacks.} In our setting, targeted
poisoning attacks are performed by adversaries against the globally
trained model. An adversary has a defined poisoning goal: increase the
probability of one class being classified as a different, incorrect
class without influencing the output probabilities of any other class.
To minimize the possibility of the poisoning attack being detected,
the prediction outcomes of classes auxiliary to the attack should be
minimally changed.

We assume that poisoning attacks are performed using either the
label-flipping strategy~\cite{Biggio:2012}, in which the labels of
honest training examples are flipped to a target class, or the
backdoor strategy~\cite{Bagdasaryan:2018, Gu:2017}, in which unused
features are augmented with secret patterns and exploited.

Since the range of possible SGD updates from an adversary is
unbounded, another possible attack involves scaling malicious
updates to overpower honest clients. However, state of the art
magnitude-based detection methods exist~\cite{Blanchard:2017} to
prevent these attacks, and therefore we do not consider these
attacks. % in our solution.

Sybils perform poisoning attacks on federated learning by providing
updates that direct the shared model towards a common poisoned
objective, as shown in Figure~\ref{fig:sgd-detail}. We do not
constrain how a sybil selects these poisoned updates: they can be
sourced from poisoned data~\cite{Biggio:2011} or computed 
through other methods.

\noindent \textbf{Attacker capability.} For a poisoning attack to
succeed in a deployment with many honest clients, an attacker must
exert more influence on their target class than the total honest
influence on this class. This is already a problem for classes that
are not well represented in the training data, but in federated
learning, where each client is given an equal share of the aggregated
gradient, attackers can attack any class with enough influence by
generating additional sybils.

If a third party account verification process exists in the system, we
assume that the adversary has the means to bypass it, either by
creating malicious accounts or by compromising honest
clients/accounts. 

Sybils observe global model state and send any arbitrary gradient
contribution to aggregator at any iteration. Sybils can collude by
sharing state among themselves and by sending updates in an
intelligent, coordinated fashion. Sybils controlled by multiple sets
of non-colluding adversaries may perform poisoning attacks
concurrently.

%!TEX root = paper.tex
%%%%%%%%%%%%%%%%%%%%%%%%%%%%%%%%%%%%%%%%%%%%%%%%%%%%%%%%%%%%%%%%%%
% \section{SGD Challenges and existing defenses}
\section{SGD Challenges and defenses}
\label{sec:design}
%%%%%%%%%%%%%%%%%%%%%%%%%%%%%%%%%%%%%%%%%%%%%%%%%%%%%%%%%%%%%%%%%%

%%%%%%%%%%%%%%%%%%%%%%%%%%%%%%%%%%%%%%%%%%%%%%%%%%%%%%%%%%%%%%%%%%%%%%%%
%\subsection{SGD challenges}
%%%%%%%%%%%%%%%%%%%%%%%%%%%%%%%%%%%%%%%%%%%%%%%%%%%%%%%%%%%%%%%%%%%%%%%%

In the traditional federated learning setting, the federated
learning service only has access to the outputs of local SGD
computations. From the aggregator's perspective, detecting sybils from
a stream of SGD iterations is difficult. If GD, and not SGD, is used
as the optimization algorithm, then detection becomes easier: in this
setting, updates are deterministic functions of data and duplicate
updates are easy to detect. The challenge with SGD is three-fold:

\begin{enumerate}[wide=1pt,labelwidth=!, labelindent=0pt, label=
\textbf{Challenge \arabic*}.]

\item Each client houses a local, unseen partition of the data
  that independently may not satisfy the global learning objective.
  When receiving a gradient from a client, it is difficult
  for the aggregator to tell whether the gradient points towards a
  malicious objective or not. We assume that FoolsGold, the system
  we design, does not possess a validation dataset. \vspace{.5em}

\item Since only a small subset of the original dataset is used in
  each iteration, the stochastic objective changes with each
  iteration. The aggregator cannot assume that updates pointing in
  sporadic directions are malicious. As well, the aggregator cannot
  assume that updates that are similar came from similar underlying
  datasets. \vspace{.5em}
  
\item As the batch size $b$ of updates decreases, the variance of
  updates contributed by clients increases. At any given iteration,
  sampling a smaller portion of the dataset results in a higher
  variance of gradient values, producing the sporadic behavior
  described above. Since adversaries can send arbitrary updates to
  the model, we cannot trust that they will conform to a specified
  batch size configuration.
\end{enumerate}

%%%%%%%%%%%%%%%%%%%%%%%%%%%%%%%%%%%%%%%%%%%%%%%%%%%%%%%%%%%%%%%%%%%%%%%%
%\subsection{Deficiency of existing defenses}
%%%%%%%%%%%%%%%%%%%%%%%%%%%%%%%%%%%%%%%%%%%%%%%%%%%%%%%%%%%%%%%%%%%%%%%%

The challenges above make existing poisoning defenses ineffective
against sybil-based attacks on federated learning, particularly in
non-IID settings.

\noindent \textbf{Multi-Krum~\cite{Blanchard:2017}} was specifically
designed to counter adversaries in federated learning. In Multi-Krum,
the top $f$ contributions to the model that are furthest from
the mean client contribution are removed from the aggregated gradient.
Multi-Krum uses the Euclidean distance to determine which gradient
contributions are removed, requires parameterization of the number of
expected adversaries, and can theoretically only withstand sybil-based
poisoning attacks of up to 33\% adversaries in the client pool.

We assume that attackers can spawn a large number of sybils, rendering
assumptions about proportions of honest clients unrealistic. As well, the mean
in the Multi-Krum process can be arbitrarily influenced by sybil
contributions. We compare FoolsGold against Multi-Krum in
Section~\ref{sec:prioreval}.

%%%%%%%%%%%%%%%%%%%%%%%%%%%%%%%%%%%%%%%%%%%%%%%%%%%%%%%%%%%%%%%%%%%%%%%%
\section{FoolsGold design}
\label{sec:designdetail}
%%%%%%%%%%%%%%%%%%%%%%%%%%%%%%%%%%%%%%%%%%%%%%%%%%%%%%%%%%%%%%%%%%%%%%%%

Our solution is intended for a federated learning setting
where the service only has access to the outputs of client SGD
computations. We design a learning method that does not make
assumptions about the proportion of honest clients in the system. Our
learning method only uses state from the learning process itself to
adapt the learning rates of clients.

Our key insight is that \emph{honest clients can be separated from
  sybils by the diversity of their gradient updates}. In federated
  learning, since each client's training data has a unique distribution
  and is not shared, sybils share a common objective and will contribute
  updates that appear more similar to each other than honest clients. This
  property of federated learning is shown in 
  Figure~\ref{fig:squeezenet_grads} and explored in 
  Section~\ref{sec:noniid}.

FoolsGold uses this assumption to modify the learning rates
of each client in each iteration. Our approach aims to maintain the
learning rate of clients that provide unique gradient updates, while
reducing the learning rate of clients that repeatedly contribute
similar-looking gradient updates.

% This is akin to state of the art
% sybil defense work which assumes that fake clients are easy to create,
% but \emph{genuine interactions and genuine data} are difficult to
% create. For example, it is difficult to forge connections with real
% users in a social network~\cite{Tran:2009}, or to forge genuine
% timestamps of user activity in an online system~\cite{Viswanath:2015}.

With this in mind, FoolsGold has five design goals:

\begin{enumerate}[wide=1pt, labelwidth=!, labelindent=0pt, label=\textbf{Goal \arabic*}.]

  \item When the system is not attacked, FoolsGold should preserve the
  performance of federated learning.
    \vspace{.5em}

  \item FoolsGold should devalue contributions from clients that
    point in similar directions.\vspace{.5em}

  \item FoolsGold should be robust to an increasing number of sybils
    in a poisoning attack. \vspace{.5em}

  \item FoolsGold should distinguish between honest updates that
    mistakenly appear malicious due to the variance of SGD and sybil
    updates that operate on a common malicious objective.
    \vspace{.5em}

  \item FoolsGold should not rely on external assumptions about the
    clients or require parameterization about the number of attackers.

\end{enumerate}

%%%%%%%%%%%%%%%%%%%%%%%%%%%%%%%%%%%%%%%%%%%%%%%%%%%%%%%%
\begin{figure}[t]
  \centering
    \includegraphics[width=.85\linewidth]{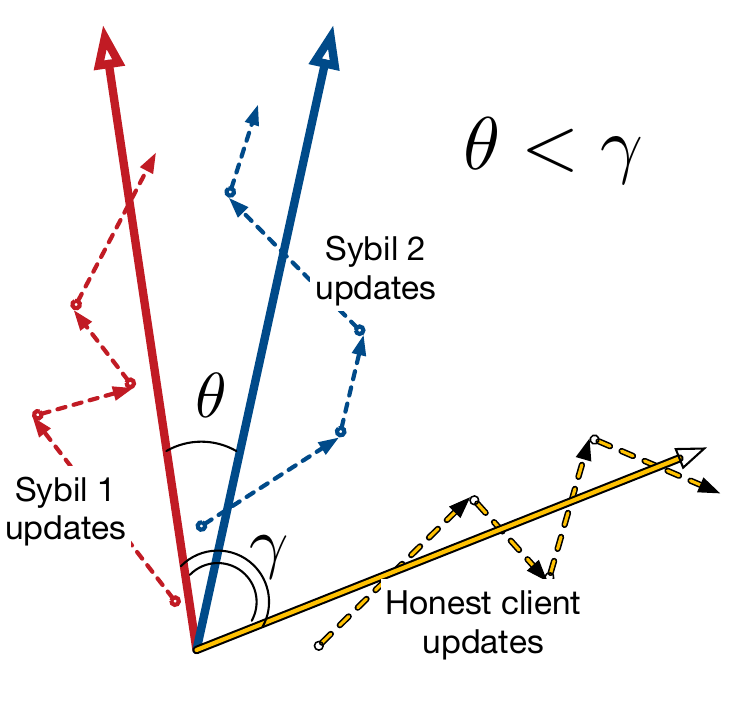}
    \caption{Dashed lines are gradient updates from three clients (2
      sybils, 1 honest). Solid lines are aggregated update vectors.
      The angle between the aggregated update vectors of sybil
      clients ($\theta$) is smaller than between those of the honest
      client and a sybil ($\gamma$). Cosine similarity would reflect
      this similarity.}
    \label{fig:agg-gradient-sim}
\end{figure}
%%%%%%%%%%%%%%%%%%%%%%%%%%%%%%%%%%%%%%%%%%%%%%%%%%%%%%%%

%%%%%%%%%%%%%%%%%%%%%%%%%%%%%%%%%%%%%%%%%%%%%%%%%%%%%%%%%%%%
\begin{algorithm}[t]
\DontPrintSemicolon
  \KwData{Global Model $w^t$ and SGD update $\Delta_{i,t}$ from each
  client $i$ at iteration $t$. Confidence parameter $\kappa$}
  \For{Iteration $t$}{
    \For{All clients $i$} {
        \tcp{Updates history}
        Let $H_i$ be the aggregate historical vector $\sum^T_{t=1}
        \Delta_{i,t}$\;
        \tcp{Feature importance}
        Let $S_t$ be the weight of indicative features at iteration
        $t$\;
        \For{All other clients $j$}{
            Let $cs_{ij}$ be the $S_t$-weighted cosine similarity
            between $H_i$ and $H_j$\;
        }
        Let $v_i = \max_j(cs_i)$ 
    }    
    
    \For{All clients $i$}{
      \For{All clients $j$}{
        \tcp{Pardoning}
        \If {$v_j > v_i$} {
          $cs_{ij} \mathrel{{*}{=}} v_i / v_j$ \;
        }
      }
      \tcp{Row-size maximums}
      Let $\alpha_i = 1 - \max_j(cs_i)$\;
    }

    \tcp{Logit function}
    $\alpha_i = \alpha_i / \max_i(\alpha)$\;
    $\alpha_i = \kappa(\ln[(\alpha_i)/(1 - \alpha_i)] + 0.5)$\;
    \tcp{Federated SGD iteration}
    $w_{t+1} = w_{t} + \sum_i \alpha_i \Delta_{i,t}$

  }
  \caption{FoolsGold algorithm.\label{alg:foolsgold}}
\end{algorithm}
%%%%%%%%%%%%%%%%%%%%%%%%%%%%%%%%%%%%%%%%%%%%%%%%%%%%%%%%%%%%

We now explain the FoolsGold approach (Algorithm~\ref{alg:foolsgold}).
In the federated learning protocol, gradient updates are collected and
aggregated in synchronous update rounds. FoolsGold adapts the learning
rate $\alpha_i$ per client\footnote{Note that we use $\alpha$ for the
FoolsGold assigned learning rate, and $\eta$ for the traditional,
local learning rate. These are independent of each other.} based on 
(1) the update similarity among indicative features in any given
iteration, and (2) historical information from past iterations.

\noindent \textbf{Cosine similarity.} We use cosine similarity to
measure the angular distance between updates. This
is preferred to Euclidean distance since sybils can manipulate
the magnitude of a gradient to achieve dissimilarity, but the
direction of a gradient cannot be manipulated without reducing attack
effectiveness. The magnitude of honest updates is also
affected by client-side hyper-parameters, such as the local learning
rate, which we do not control.

\noindent \textbf{Feature importance.} 
From the perspective of a potential poisoning attack, there are three
types of features in the model: (1) features that are relevant to
the correctness of the  model, but must be modified for a successful
attack, (2) features that are relevant to the correctness of the
model, but irrelevant for the attack, and (3) features that are
irrelevant to both the attack and the model.

Similar to other decentralized poisoning defenses~\cite{Shen:2016}, we
look for similarity only in the indicative features (type 1 and 2) in
the model. This prevents adversaries from manipulating irrelevant
features while performing an attack, which is evaluated in
Section~\ref{sec:featurenoise}.

The indicative features are found by measuring the magnitude of
model parameters in the output layer of the global model. Since training
data features and gradient updates are normalized while performing SGD, the
magnitude of an output layer parameter maps directly to its influence on
the prediction probability~\cite{Shokri:2015}. These features can be
filtered (hard) or re-weighed (soft) based on their influence on the
model, and are normalized across all classes to avoid biasing one
class over another.

For deep neural networks, we do not consider the magnitude of values
in the non-output layers of the model, which do not map directly to output
probabilities and are more difficult to reason about. Recent work on
feature influence in deep neural networks~\cite{Ancona:2017, Datta:2016}
may better capture the intent of sybil-based poisoning attacks in deep
learning and we leave this analysis as future work.

%% At each iteration, we maintain a history of the cosine similarity
%% between two clients.
%% %
%% \ivan{I thought that we only maintained the aggregate vector for each
%%   client, not pair-wise similarity between clients. Is this text out of
%%   date?}
%% %
%% This is done by maintaining a total sum of a
%% clients gradient contributions, and evaluating the cosine similarity of
%% each pair of clients.
%% %
%% \ivan{cannot eval cos sim over clients; has to be over their
%%   data.. but what data? rewrite/clarify.}
%% %
%% We use the similarity on an aggregated gradient
%% instead of the similarity from an isolated iteration to better estimate
%% the expectation of the gradient direction similarities between two
%% clients. 

\noindent \textbf{Updates history.}
FoolsGold maintains a history of updates from each client. It
does this by aggregating the updates at each iteration from a single
client into a single aggregated client gradient (line 3). To better
estimate similarity of the overall contributions made by clients,
FoolsGold computes the similarity between pairwise aggregated
historical updates instead of just the updates from the current
iteration.

Figure~\ref{fig:agg-gradient-sim} shows that even for two sybils with
a common target objective, updates at a given iteration may diverge
due to the problems mentioned in Challenge 2. However, the cosine
similarity between the sybils' aggregated historical updates is
high, satisfying Goal 2. 

We interpret the cosine similarity on the indicative features, a value
between -1 and 1, as a representation of how strongly two clients are
acting as sybils. We define $v_i$ as the maximum pairwise similarity
for a client $i$, ensuring that as long as one such interaction
exists, we can devalue the contribution while staying robust to an
increasing number of sybils, as prescribed by Goal 3.
\begin{align*}
    cs_{ij} = cosine\_similarity(\sum^T_{t=1} \Delta_{i,t}, 
    \sum^T_{t=1} \Delta_{j,t})
\end{align*}

\noindent \textbf{Pardoning.}
Since we have weak guarantees on the cosine similarities between an
honest client and sybils, honest clients may be incorrectly penalized
under this scheme. We introduce a pardoning mechanism that avoids
penalizing such honest clients by re-weighing the cosine similarity by
the ratio of $v_i$ and $v_j$ (line 13), satisfying Goal 4. The new
client learning rate $\alpha_i$ is then found by inverting the
maximum similarity scores along the 0-1 domain. Since we assume at
least one client in the system is honest, we rescale the vector such
that the maximum adaption of the learning rate is 1 (line 18).
This ensures that at least one client will have an unmodified update
and encourages the system towards Goal 1: a system containing only
honest nodes will not penalize their contributions. 
\begin{align*}
  \alpha_i &= 1 - \max_j(cs_i) \\
  \alpha_i &= \frac{\alpha_i}{\max_i(\alpha)}
\end{align*}

\noindent \textbf{Logit.} However, even for very similar updates, the
cosine similarity may be less than one. An attacker may exploit this
by increasing the number of sybils to remain influential. We therefore
want to encourage a higher divergence for values that are near the two
tails of this function, and avoid penalizing honest clients with a
low, non-zero similarity value. Thus, we use the logit function (the
inverse sigmoid function) centered at 0.5 (line 19), for these
properties. We also expose a confidence parameter $\kappa$ that scales
the logit function and show in~\ref{sec:converge} that
$\kappa$ can be set as a function of the data distribution among
clients to guarantee convergence.
\begin{align*}
    \alpha_i &= \kappa(\ln[\frac{\alpha_i}{1 - \alpha_i}] + 0.5)
\end{align*}

When taking the result of the logit function, any value exceeding
the 0-1 range is clipped and rounded to its respective boundary
value. Finally, the overall gradient update is calculated by applying
the final re-scaled learning rate:
\begin{align*}
    w_{t+1} = w_t + \sum_i \alpha_{i} \Delta_{i,t} 
\end{align*}

Note that this design does not require parameterization of the
expected number of sybils or their properties (Goal 5), is
independent of the underlying model, and is also independent of SGD
details such as the local client's learning rate or batch size.

\noindent \textbf{Augmenting FoolsGold with other methods.}
Simply modifying the learning rates of clients based on their
aggregate gradient similarity will not handle all poisoning attacks.
Clearly, an attack from a single adversary will not exhibit such
similarity. FoolsGold is best used when augmented with
existing solutions that detect poisoning attacks from a bounded number
of attackers. We evaluate FoolsGold with Multi-Krum
in Section~\ref{sec:combined}. 

\noindent \textbf{Convergence properties.}
FoolsGold is analagous to importance sampling~\cite{Needell:2014} and
adaptive learning rate methods~\cite{Zeiler:2012}, which have both been
applied to SGD algorithms and have convergence guarantees.
We analyze FoolsGold's convergence guarantees in~\ref{sec:converge}.
Our experimental results further support that FoolsGold converges
under a variety of conditions.

\noindent \textbf{FoolsGold security guarantees.} 
We claim that our design mitigates an adversary performing a
targeted poisoning attack by limiting the influence they gain through
sybils. We also claim that
FoolsGold satisfies the specified design goals: it
preserves the updates of honest nodes while penalizing the
contributions of sybils. In the next section we empirically validate
these claims across several different dimensions using a prototype of
FoolsGold.

%!TEX root = paper.tex
%%%%%%%%%%%%%%%%%%%%%%%%%%%%%%%%%%%%%%%%%%%%%%%%%%%%%%%%%%%%%%%%%%
\section{Evaluation}
\label{sec:eval}
%%%%%%%%%%%%%%%%%%%%%%%%%%%%%%%%%%%%%%%%%%%%%%%%%%%%%%%%%%%%%%%%%%

We evaluate FoolsGold by implementing a federated learning prototype
in 600 lines of Python. The prototype includes 150 lines for
FoolsGold, implementing Algorithm~\ref{alg:foolsgold}. We use
scikit-learn~\cite{scikit-learn} to compute cosine similarity of
vectors. For each experiment below, we partition the original training
data into disjoint non-IID training sets, locally compute SGD updates
on each dataset, and aggregate the updates using the described
FoolsGold method to train a globally shared classifier.

We evaluate our prototype on four well-known classification datasets:
MNIST~\cite{Lecun:1998}, a digit classification problem, 
VGGFace2~\cite{Cao:2018}, a facial recognition problem, 
KDDCup~\cite{Dua:2017}, which contains classified network intrusion
patterns, and Amazon~\cite{Dua:2017}, which contains a corpus of text from
product reviews. Table~\ref{tab:datasets} describes each dataset.
\newcolumntype{R}[1]{>{\raggedleft\let\newline\\\arraybackslash\hspace{0pt}}m{#1}}

%%%%%%%%%%%%%%%%%%%%%%%%%%%%%%%%%%%%%%%%%%%%%%%%%%%%%%%%%%%%%%%%%%%%%%%%
\begin{table}[t]
\caption{Datasets used in this evaluation.}
\label{tab:datasets}
\centering
\begin{tabular}{ r| R{1.1cm}R{1.05cm}R{1.3cm}p{2cm} }
 \hline
 \textbf{Dataset} & \textbf{Train Set Size} & \textbf{Classes} & 
 \textbf{Features} & \textbf{Model Used} \\
 \hline
 MNIST & 60,000 & 10 & 784 & 1-layer softmax \\
 \hline
 VGGFace2 & 7,380 & 10 & 150,528 & SqueezeNet, VGGNet11 \\
 \hline
 KDDCup & 494,020 & 23 & 41 & 1-layer softmax \\ 
 \hline
 Amazon & 1,500 & 50 & 10,000 & 1-layer softmax \\
 \hline
 \end{tabular} 
\end{table}
%%%%%%%%%%%%%%%%%%%%%%%%%%%%%%%%%%%%%%%%%%%%%%%%%%%%%%%%%%%%%%%%%%%%%%%%

Each dataset was selected for one of its particularities. MNIST was
chosen as the baseline dataset for evaluation since it was used
extensively in the original federated learning evaluation~\cite{McMahan:2017}.
The VGGFace2 dataset was chosen as a more complex learning task that
required deep neural networks to solve. For simplicity in evaluating
poisoning attacks, we limit this dataset to the top 10 most frequent classes
only. The KDDCup dataset has a relatively low number of features, and
contains a massive class imbalance: some classes have as few as 5 examples,
while some have over 280,000. Lastly, the Amazon dataset is unique in that
it has few examples and contains text data: each review is converted into
its one hot encoding, resulting in a large feature vector of size 10,000. 

For all the experiments in this section, targeted poisoning attacks
are performed that attempt to encourage a \emph{source label} to be
classified as a \emph{target label} while training on a
federated learning prototype. When dividing the data, each class is
always completely represented by a single client, which is
consistent with the federated learning baseline. In
all experiments the number of honest clients in the system varies by
dataset: 10 for MNIST and VGGFace2, 23 for KDDCup, and 50 for Amazon. We
consider more IID settings in Section~\ref{sec:noniid}.

For MNIST, KDDCup, and Amazon, we train a single layer fully-connected
softmax for classification. The size of the trained model in each case is
the product of the number of classes and number of features. For
VGGFace2, we use two popular architectures pre-trained on Imagenet
from the torchvision package~\cite{Marcel:2010}: SqueezeNet1.1, 
a smaller model designed for edge devices with 727,000 parameters; and
VGGNet11, a larger model with 128 million parameters. When using FoolsGold
to compare client similarity, we only use the features in the final output
layer's gradients (fully connected layer in VGGNet and 10 1x1 convolutional
kernels in SqueezeNet). 

In MNIST, the data is already divided into 60,000 training
examples and 10,000 test examples~\cite{Lecun:1998}. For VGGFace2, KDDCup
and Amazon, we randomly partition 30\% of the total data to be test data.
The test data is used to evaluate two metrics that represent the
performance of our algorithm: the \emph{attack rate}, which is the
proportion of attack targets (source labels for label flipping
attacks or embedded images for backdoor attacks) that are incorrectly
classified as the target label, and the \emph{accuracy}, which is the
proportion of examples in the test set that are correctly classified. 

The MNIST and KDDCup datasets were executed with 3,000 iterations
and a batch size of 50 unless otherwise stated. For Amazon, due to the
high number of features and low number of samples per class, we train
for 100 iterations and a batch size of 10. For VGGFace2, we train for 500
iterations, using SGD with batch size of 8, momentum 0.9,
weight decay 0.0001, and learning rate 0.001. These % model parameterization
values were found using cross validation in the training set. During
training, images were resized to 256x256 and randomly cropped and flipped
to 224x224. During testing, images were resized to 256x256 and a 224x224
center was cropped.

In each of the non-attack scenarios, we ran each experiment to
convergence. In all attack scenarios, we found that our selected number of
iterations was sufficiently high such that the performance of the attack
changed minimally with each iteration, indicating the result of the attack
to be consistent. For each experiment, FoolsGold is parameterized
with a confidence parameter $\kappa = 1$, and does not use the
historical gradient or the significant features filter (we evaluate
these design elements independently in Section~\ref{sec:evalmodule} 
and~\ref{sec:featurenoise}, respectively). Each reported data point
is the average of 5 experiments.

%%%%%%%%%%%%%%%%%%%%%%%%%%%%%%%%%%%%%%%%%%%%%%%%%%%%%%%%%%%%
\subsection{Canonical attack scenarios}
%%%%%%%%%%%%%%%%%%%%%%%%%%%%%%%%%%%%%%%%%%%%%%%%%%%%%%%%%%%%

Our evaluation uses a set of 5 attack scenarios (that we term
canonical for this evaluation) across the three datasets
(Table~\ref{tab:attackeval}). Attack \textbf{A-1} is a traditional
poisoning attack: a single client joins the federated learning system
with poisoned data. Attack \textbf{A-5} is the same attack
performed with 5 sybil clients joining the system. Each client sends
updates for a subset of its data through SGD, meaning that their
updates are not identical. Attack \textbf{A-2x5} evaluates FoolsGold's
ability to thwart multiple attacks at once: 2 sets of client sybils attack
the system concurrently, and for attack evaluation purposes we assume that
the classes in these attacks do not overlap\footnote{We do not perform a
1-2 attack in parallel with a 2-3 attack, since evaluating the 1-2 attack
would be biased by the performance of the 2-3 attack.}.

Since KDDCup99 is a unique dataset with severe class imbalance,
instead of using an A-2x5 attack we choose to perform a different
attack, \textbf{A-OnOne}, on this dataset. In KDDCup99, data from
various network traffic patterns are provided. Class ``Normal''
identifies patterns without any network attack, and is proportionally
large (approximately 20\% of the data) in the overall dataset. 
Therefore, when attacking KDDCup99, we assume that adversaries mis-label
malicious attack patterns, which are proportionally small, 
(approximately 2\% of the data) and poison the malicious class towards
the \say{Normal} class. A-AllOnOne is a unique attack for KDDCup in
which 5 different malicious patterns are each labeled as \say{Normal},
and each attack is performed concurrently.

Finally, we use \textbf{A-99} to illustrate the robustness of
FoolsGold to massive attack scenarios. In this attack, an adversary
generates 990 sybils to overpower a network of 10 honest
clients and all of them attempt a single 1-7 attack against MNIST.  

\begin{table}[t]
\caption{Canonical attacks used in our evaluation.}
\label{tab:attackeval}
\centering
\begin{tabular}{ r | p{3.5cm}  p{2cm}}
 \hline
 \textbf{Attack} & \textbf{Description} & \textbf{Dataset} \\
 \hline
 A-1 & Single client attack. & All \\
 \hline
 A-5 & 5 clients attack. & All \\
 \hline
 % A-2x5 & 2 sets of 5 clients, concurrent attacks. &
 % MNIST, Amazon, VGGFaces2 \\
 % \hline
 A-5x5 & 5 sets of 5 clients, concurrent attacks. &
 MNIST,\hspace{1cm} Amazon, VGGFaces2 \\
 \hline
 A-OnOne & 5 clients executing 5 attacks on the same target
 class. & KDDCup99 \\
 \hline
 A-99 & 99\% sybils, performing the same attack. &
 MNIST \\
 \hline
 \end{tabular} 
\end{table}

%%%%%%%%%%%%%%%%%%%%%%%%%%%%%%%%%%%%%%%%%%%%%%%%%%%%%%%%%%%%%%%%%%%%%%%%

\begin{figure*}[t]
    \includegraphics[width=\linewidth]{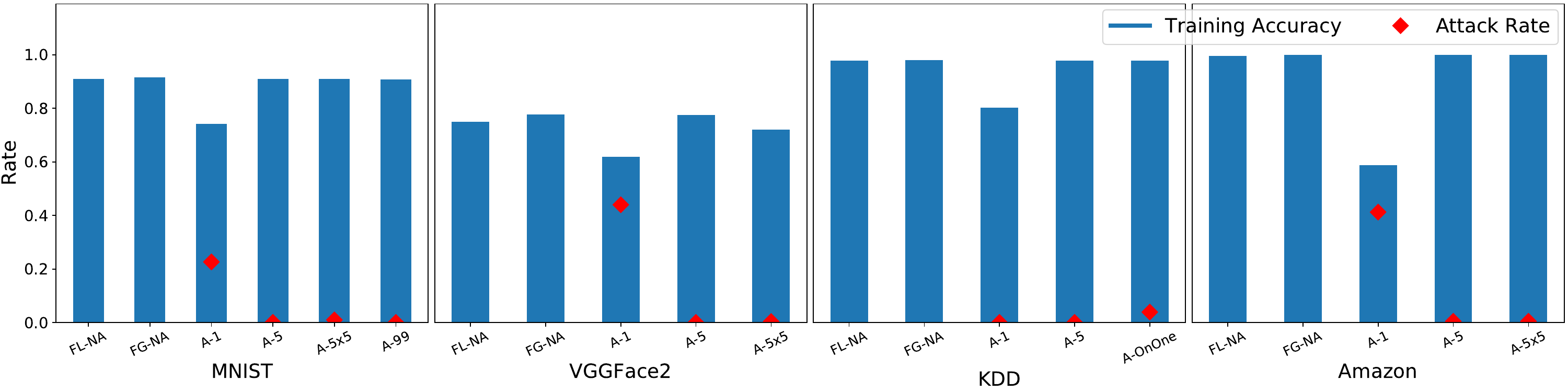}
    \caption{Training accuracy (blue bars) and attack rate (red ticks) for
      canonical attacks against the relevant canonical datasets.
    }
    \label{fig:canon-multi}
\end{figure*}

%%%%%%%%%%%%%%%%%%%%%%%%%%%%%%%%%%%%%%%%%%%%%%%%%%%%%%%%%%%%%%%%%%%%%%%%

Since we use these canonical attacks throughout this work, we first
evaluate each attack on the respective dataset
(Table~\ref{tab:attackeval}) with FoolsGold enabled.
Figure~\ref{fig:canon-multi} plots the attack rate and
test accuracy for each attack in Table~\ref{tab:attackeval}. The
Figure also shows results for the system without attacks: the original
federated learning algorithm (FL-NA) and the system with
the FoolsGold algorithm (FG-NA).

Figure~\ref{fig:canon-multi} shows that for most attacks, including the A-OnOne attack and the
A-99 attack, FoolsGold effectively prevents the attack while
maintaining high training accuracy. As FoolsGold faces larger groups
of sybils, it has more information to more reliably detect
similarity between sybils. FoolsGold performs worst on the
A-1 attacks in which only one malicious client attacked the system.
The reason is simple: without multiple colluding sybils, malicious
and honest clients are indistinguishable to the FoolsGold aggregator.

Another point of interest is the prevalence of false positives. In
A-1 KDDCup, our system incorrectly penalized an honest client
for colluding with the attacker, lowering the prediction rate of the
honest client as the defense was applied. We observe that the two
primary reasons for low training error are either a high
attack rate (false negatives) or a high target class error rate (false
positives). We also discuss false positives from data similarity in 
Section~\ref{sec:noniid}.

%%%%%%%%%%%%%%%%%%%%%%%%%%%%%%%%%%%%%%%%%%%%%%%%%%%%%%%%%%%%
\subsection{Comparison to prior work}
\label{sec:prioreval}
%%%%%%%%%%%%%%%%%%%%%%%%%%%%%%%%%%%%%%%%%%%%%%%%%%%%%%%%%%%%

We compare FoolsGold to Multi-Krum aggregation~\cite{Blanchard:2017}, the
known state of the art in defending against Byzantine clients in
distributed SGD (see Section~\ref{sec:background}).

\noindent\textbf{Comparison to Multi-Krum.} In this experiment we
compare FoolsGold to Multi-Krum and an unmodified federated learning
baseline as we vary the number of sybils.

We implemented Multi-Krum as specified in the original 
paper~\cite{Blanchard:2017}: at each iteration, the total Euclidean
distance from the $n - f - 2$ nearest neighbors is calculated for
each update. The $f$ updates with the highest distances are removed
and the average of the remaining updates is calculated. Multi-Krum
relies on the $f$ parameter: the maximum number of Byzantine clients
tolerated by the system. To defend against sybil attacks in this
setting, we set $f$ to be the number of sybils executing a poisoning
attack. Although this is the best case for Multi-Krum, we note that
prior knowledge of $f$ is an unrealistic assumption when
defending against sybils.

While running Multi-Krum, we found that the performance was especially
poor in the non-IID setting. When the variance of updates among
honest clients is high, and the variance of updates among sybils is
lower, Multi-Krum removes honest clients from the system. This makes
Multi-Krum unsuitable for defending against sybils in federated
learning, which is intended for non-IID 
settings~\cite{McMahan:2017}.

\begin{figure}[t]
    \includegraphics[width=\linewidth]{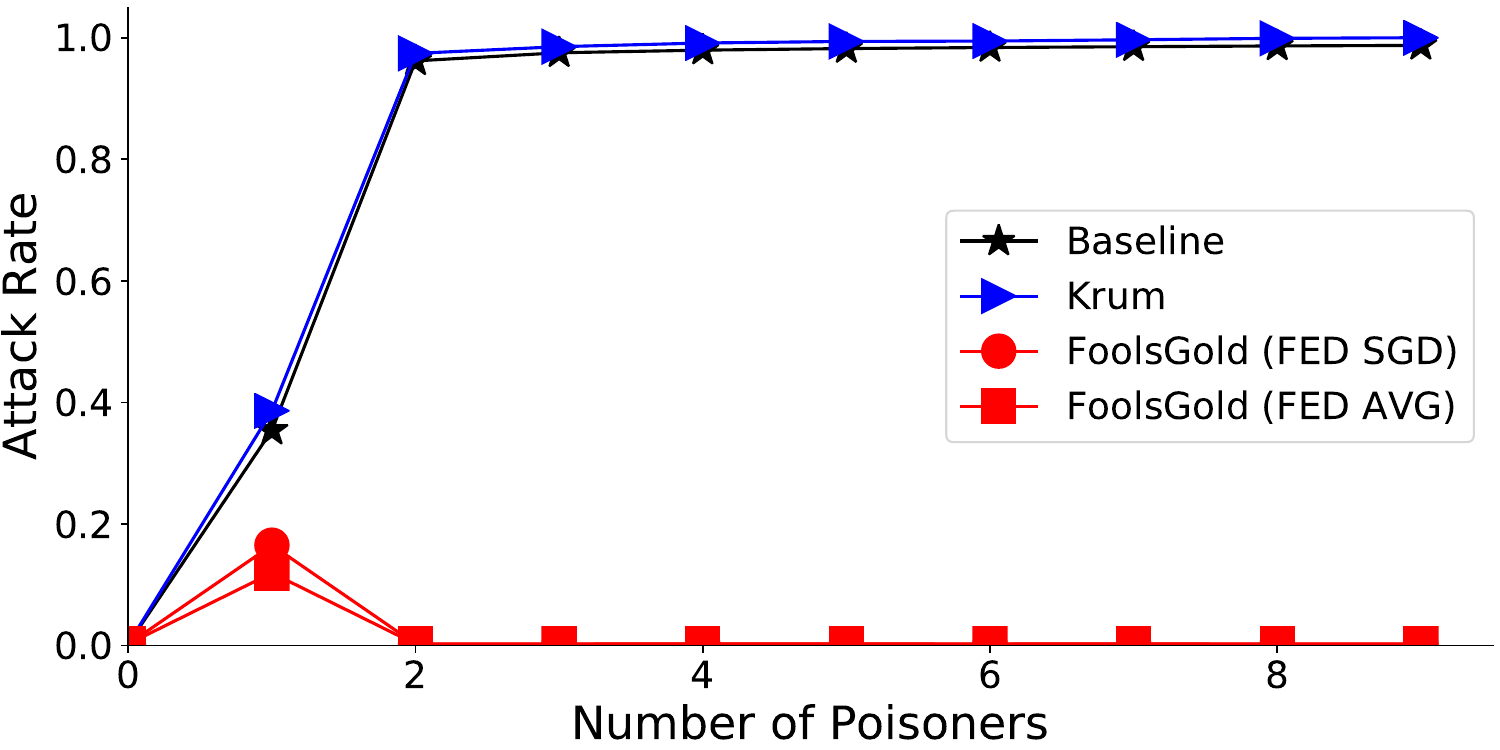}
    \caption{Attack rate for varying number of sybils, for
      federated learning (Baseline), Multi-Krum, and FoolsGold.}
    \label{fig:baselines}
\end{figure}

Figure~\ref{fig:baselines} shows the performance of the three
approaches against an increasing number of poisoners: a 1-7 attack is
performed on an unmodified non-IID federated learning system 
(Baseline), a federated learning system with Multi-Krum aggregation,
and our proposed solution. We show FoolsGold's effectiveness against
both FEDSGD and FEDAVG~\cite{McMahan:2017}, in which clients perform
multiple local iterations before sharing updates with the aggregator.

We see that as soon as the proportion of poisoners for a single class
increases beyond the corresponding number of honest clients that hold
that class (which is 1), the attack rate increases significantly for
naive averaging (Baseline). 

In addition to exceeding the parameterized number of expected
sybils, an adversary can also influence the mean client
contribution at any given iteration, and Multi-Krum will fail to
distinguish between honest clients and sybils.  

Multi-Krum works with up to 33\% sybils~\cite{Blanchard:2017}, but
fails above this threshold. By contrast, FoolsGold penalizes attackers
further as the proportion of sybils increases, and in this
scenario FoolsGold remains robust even with 9 attackers.

Consistent with the results in Figure~\ref{fig:canon-multi}, 
FoolsGold in Figure~\ref{fig:baselines} performs the worst when only one
poisoner is present. We also show that FoolsGold similarly outperforms
Multi-Krum in defending against backdoor attacks~\cite{Gu:2017} in
% Appendix~\ref{sec:extraeval}.
Appendix~B.

%%%%%%%%%%%%%%%%%%%%%%%%%%%%%%%%%%%%%%%%%%%%%%%%%%%%%%%%%%%%%%%%%%%%%%%%
%%%%%%%%%%%%%%%%%%%%%%%%%%%%%%%%%%%%%%%%%%%%%%%%%%%%%%%%%%%%%%%%%%%%%%%%

%%%%%%%%%%%%%%%%%%%%%%%%%%%%%%%%%%%%%%
\subsection{Varying client data distributions}
\label{sec:noniid}
%%%%%%%%%%%%%%%%%%%%%%%%%%%%%%%%%%%%%%

FoolsGold relies on the assumption that training data is sufficiently
dissimilar between clients. However, a realistic scenario may involve
settings where each client has overlapping data, such as images of the same people. 

To test FoolsGold under diverse data distribution assumptions, we conduct
an A-5 0-1 attack (with 5 sybils) on MNIST and VGGFace2 with 10 honest clients while
varying the proportion of shared data within both the sybils and honest
clients. We varied the distribution of sybils and honest clients over a
grid ($x_{sybil}, x_{honest} \in \{0, 0.25, 0.5, 0.75, 1\}$), where $x$
refers to the ratio of disjoint data to shared data. Specifically, a
proportion of $0\%$ refers to a non-IID setting in which each client's dataset
is composed of a single class, and $100\%$ refers to a setting where each
client's dataset is uniformly sampled from all classes. A proportion of $x$
refers to a setting where clients hold $x\%$ uniform data from all classes
and $(1 - x)\%$ disjoint data for a specific class. To create a sybil
client with $x\%$ shared data, we first create an honest client with $x\%$
shared data and $(1 - x)\%$ disjoint data using class 0, and then flip all
0 labels to a 1 to perform a targeted attack.

\begin{figure}[t]
  \includegraphics[width=\linewidth]{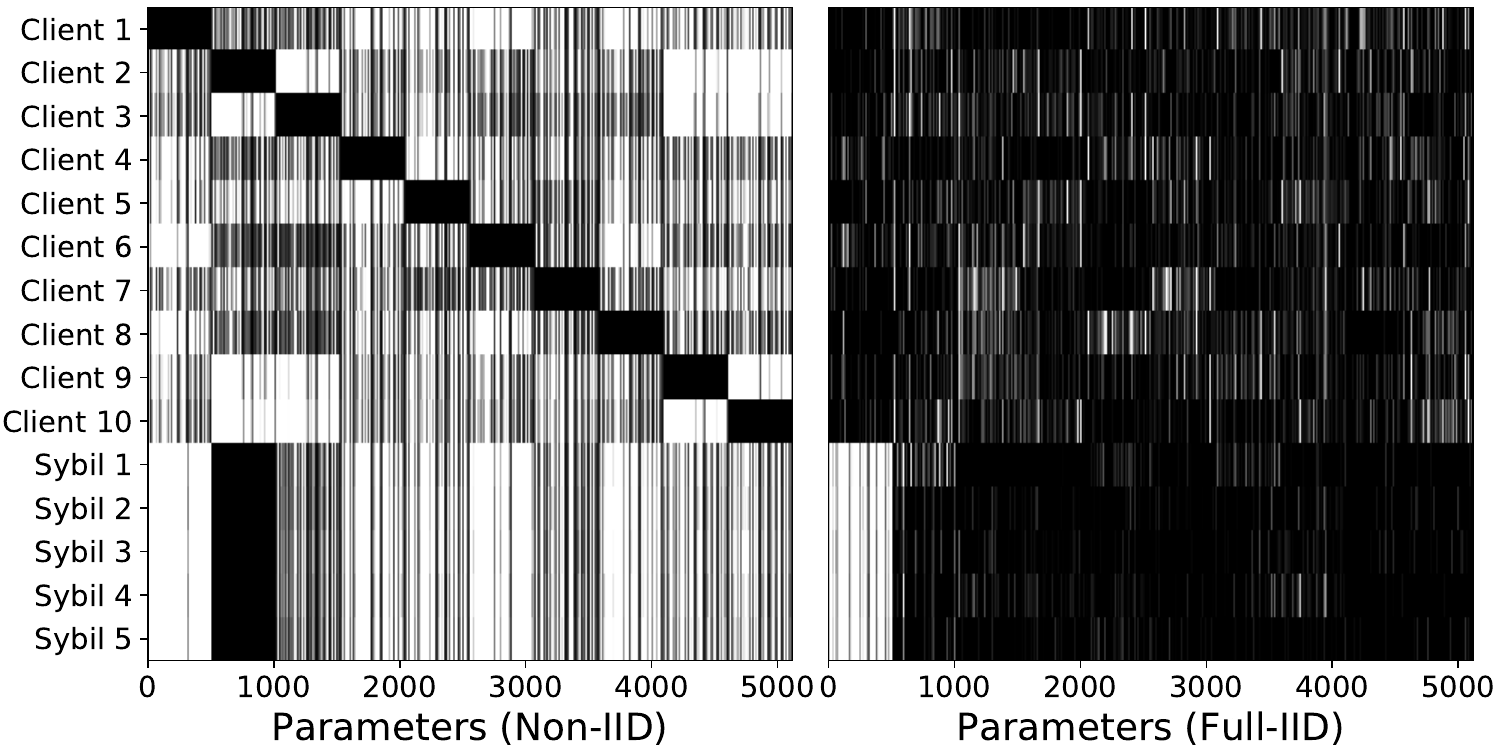}
  \caption{
    A visualization of the historical gradient of clients in Squeezenet,
    in the non-IID (left) and IID (right)
    case. The top 10 rows show the gradient vector of honest clients; the
    bottom 5 rows for sybils performing a targeted 0-1 attack.
  }
  \label{fig:squeezenet_grads}
\end{figure}

To understand how FoolsGold handles these situations,
we visualize the gradient history of all clients for a VGGFaces2 A-5
experiment in Figure~\ref{fig:squeezenet_grads}. Each
row shows the historical gradient of a client in the final classification
layer, where the top 10 rows correspond to honest clients and the bottom 5 rows correspond to
sybils. The left side of Figure~\ref{fig:squeezenet_grads} shows the
historical gradient in a non-IID setting ($x_{sybil}, x_{honest} = 0$).
Since each honest client has data corresponding to a unique class, their
gradients are highly dissimilar, and since the sybils have the same
poisoning dataset for a 0-1 attack, their gradients are highly similar.
Therefore, FoolsGold can easily detect the sybil gradients.

The right side of Figure~\ref{fig:squeezenet_grads} shows the historical gradient
in an IID setting ($x_{sybil}, x_{honest} = 1$). Even when the honest
clients all hold uniform samples of the training data, the stochasticity of
SGD introduces variance between honest clients, and the sybil gradients
are still uniquely distinct as a result of their poisoned data, as seen in
the bottom left corners of Figure~\ref{fig:squeezenet_grads}. Sybils that
intend to execute a targeted poisoning attack produce fundamentally
similar gradient updates, enabling FoolsGold to succeed. We formally define
the convergence of FoolsGold by bounds on inter-client update similarity in
%Appendix~\ref{sec:converge}.
Appendix~A.

\begin{figure}[t]
    \includegraphics[width=\linewidth]{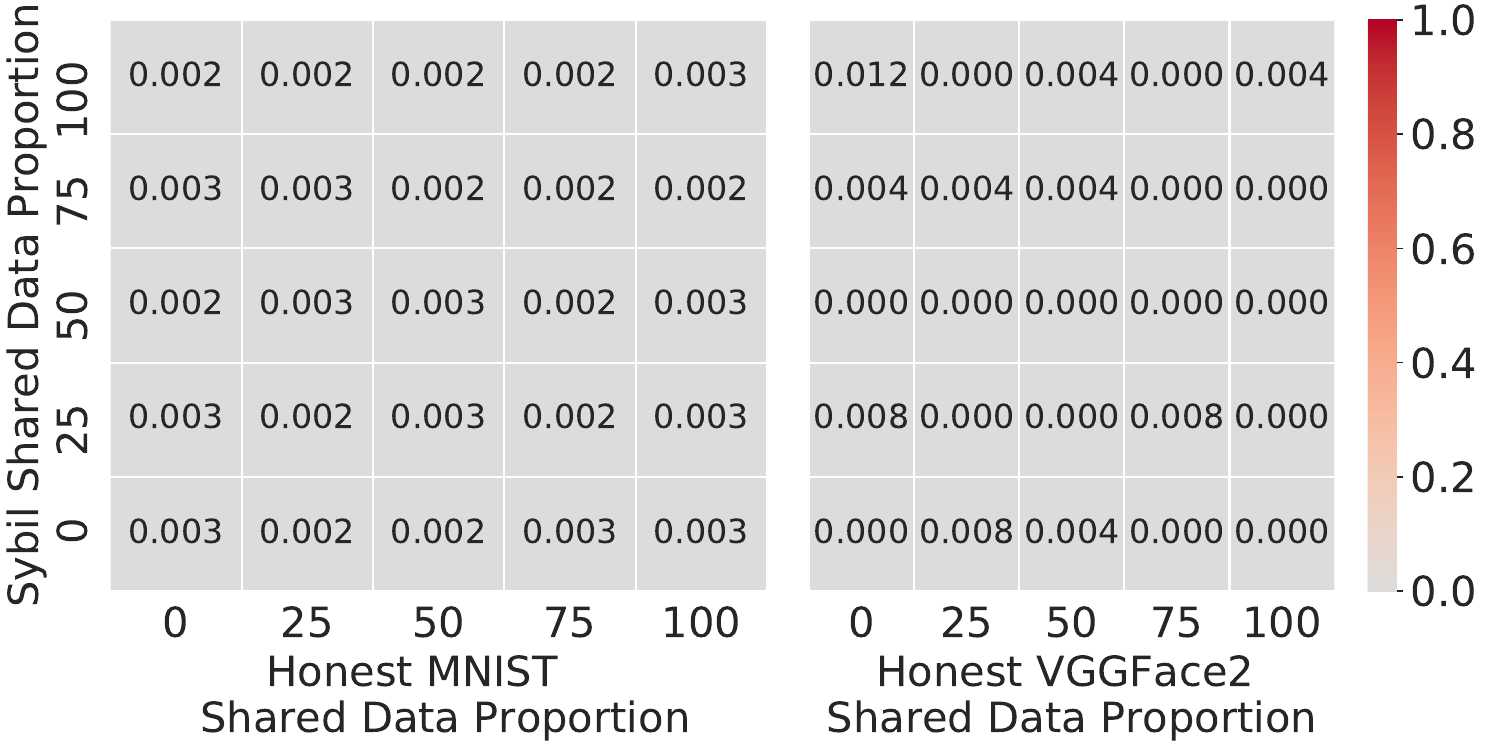}
    \caption{The attack rates on the MNIST and VGGFace2 dataset for varying
    sybil and honest client data distributions. 0\% means that the data is
    completely disjoint by class; 100\% means that the data is uniformly
    distributed.}
    \label{fig:heatmapiid}
\end{figure}

FoolsGold is effective at defending against poisoning attacks for all
combinations: the maximum attack rate was less than $1\%$ for
both the MNIST and VGGFace2 datasets, using both SqueezeNet and VGGNet.
We show these results in Figure~\ref{fig:heatmapiid}.
As the results in this section demonstrate, a malicious actor cannot
subvert FoolsGold by manipulating their malicious data
distribution. Instead, they must directly manipulate their gradient
outputs, which we explore next.

%%%%%%%%%%%%%%%%%%%%%%%%%%%%%%%%%%%%%%
\subsection{What if the attacker knows FoolsGold?}
\label{sec:subversion}
%%%%%%%%%%%%%%%%%%%%%%%%%%%%%%%%%%%%%%

If an attacker is aware of the FoolsGold algorithm, they may attempt
to send updates in ways that encourage additional dissimilarity.
This is an active trade-off: as attacker updates become less similar
to each other (lower chance of detection), they become less focused
towards the poisoning objective (lower attack utility).

We consider and evaluate four ways in which attackers may attempt to
subvert FoolsGold: (1) mixing malicious and correct data, (2) changing
sybils' training batch size, (3) perturbing contributed updates with
noise, and (4) infrequently and adaptively sending poisoned updates.
Due to space constraints, the results of (1) and (2) are listed in 
% Appendix~\ref{sec:extraeval}.
Appendix~B.

\noindent\textbf{Adding intelligent noise to updates.}
\label{sec:featurenoise}
A set of intelligent sybils could send
pairs of updates with carefully perturbed noise that is designed to
sum to zero. For example, if an attacker draws a random noise vector
$\zeta$, two malicious updates $a_1$ and $a_2$ could be contributed
as $v_1$ and $v_2$, such that: 
$v_1 = a_1 + \zeta$, and $v_2 = a_2 - \zeta$.

Since the noise vector $\zeta$ has nothing to do with the poisoning
objective, its inclusion will add dissimilarity to the malicious
updates and decrease FoolsGold's effectiveness in detecting them.
Also note that the sum of these two updates is still the same: $v_1
+ v_2 = a_1 + a_2$. This strategy can also be scaled beyond 2 sybils
by taking orthogonal noise vectors and their negation: for any subset
of these vectors, the cosine similarity is 0 or -1, while the sum
remains 0.

As explained in Section~\ref{sec:designdetail}, this attack is most
effective if $\zeta$ is only applied to features of type (3): those
which are not important for the model or the attack. This increases
sybil dissimilarity and does not adversely impact the attack.

This attack is mitigated by filtering for indicative features in the
model. Instead of looking at the cosine similarity between updates
across all features in the model, we look at a weighted cosine
similarity based on feature importance.

To evaluate the importance of this mechanism to the poisoning attack,
we execute the intelligent noise attack described above on MNIST, which
contains several irrelevant features (the large black regions) in each
example: a pair of sybils send $v_1$ and $v_2$ with intelligent noise
$\zeta$. We then vary the proportion of model parameters that are defined
as indicative from 0.001 (8 features on MNIST) to 1 (all features).

\begin{figure}[t]
    \includegraphics[width=\linewidth]{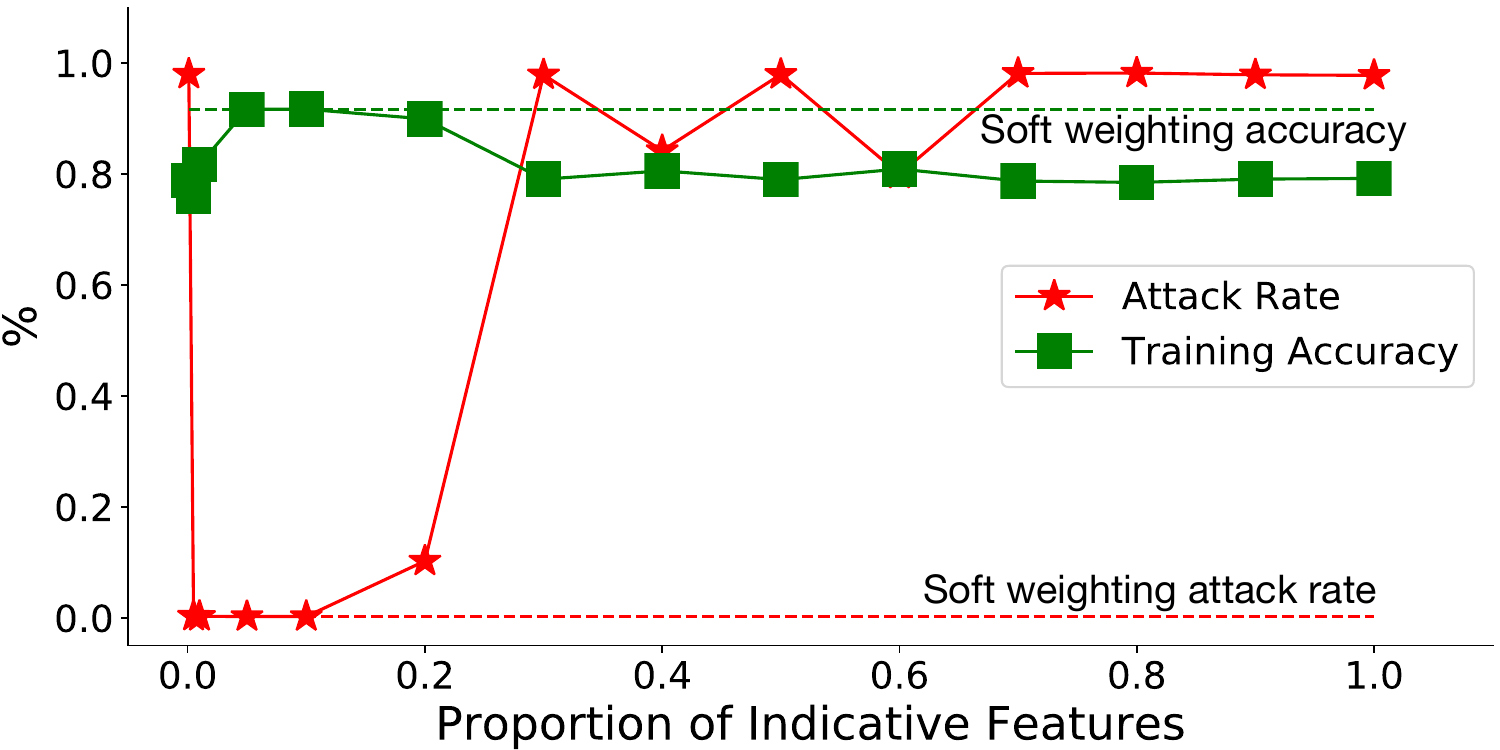}
    \caption{The performance of the optimal noisy attack on MNIST
    with varying indicative feature ratio.} 
    \label{fig:featureimportance}
\end{figure}

Figure~\ref{fig:featureimportance} shows the attack rate and the
training accuracy for varying proportions of indicative features. We
first observe that when using all of the features for similarity (far
right), the poisoning attack is successful. 

Once the proportion of indicative features decreases below 0.1 
(10\%), the dissimilarity caused by the intelligent noise is
removed from the cosine similarity and the poisoning vector dominates
the similarity, causing the intelligent noise strategy to fail with an
attack rate of near 0. We also observe that if the proportion of
indicative features is too low (0.01), the training accuracy also
begins to suffer. When considering such a low number of features,
honest clients appear to collude as well, causing false positives.

We also evaluated the soft feature weighing mechanism, which weighs
each contribution proportionally based on the model parameter itself.
The results of the soft weighting method on the same intelligent MNIST
poisoning attack are also shown in Figure~\ref{fig:featureimportance}.
For both the attack rate and training accuracy, the performance of the
soft mechanism is comparable to the optimal hard filtering mechanism.

%%%%%%%%%%%%%%%%%%%%%%%%%%%%%%%%%%%%%%%%%%%%%%%%%%%%%%%%%%%%%%%%%%%%%%%%
%%%%%%%%%%%%%%%%%%%%%%%%%%%%%%%%%%%%%%%%%%%%%%%%%%%%%%%%%%%%%%%%%%%%%%%%

\noindent\textbf{Adaptive updates method.}
We devised another optimal attack against FoolsGold that manipulates
its memory component. If an adversary knows that
FoolsGold employs cosine similarity on the update history, and is
able to locally compute the pairwise cosine similarity among sybils,
they can bookkeep this information and decide to send poisoned
updates only when their similarity is sufficiently low
(Algorithm~\ref{alg:adaptive}). This algorithm uses a parameter $M$,
representing the inter-sybil cosine similarity threshold. When $M$ is
low, sybils are less likely to be detected by FoolsGold as they will
send their updates less often; however, this will also lower the
influence each sybil has on the global model.

An adversary could generate an exceedingly large number of sybils for
a successful attack, but given that the adversary is uncertain about
the influence needed to overpower the honest clients, this
becomes a difficult trade-off to navigate for an optimal attack.

%%%%%%%%%%%%%%%%%%%%%%%%%%%%%%%%%%%%%%%%%%%%%%%%%%%%%%%%%%%%
\begin{algorithm}[t]
\DontPrintSemicolon
  \KwData{Desired $\Delta_{t}$ from all sybils at iteration $t$,
  similarity threshold $M$}
  \For{iteration $t$}{
    \For{All sybils $i$} {
        \tcp{Compute local history}
        Let $H_i$ be the aggregate historical vector $\sum^T_{t=1}
        \Delta_{i,t}$\;
        \tcp{Feature Importance and similarity}
        Let $S_t$ be the set of indicative features at iteration $t$ 
        (if known)\;
        \For{All other sybils j}{
            Let $cs_{ij}$ be the $S_t$-weighted cosine similarity
            between $H_i$ and $H_j$\;\;
        }
    }
    \For{All sybils $i$} {
        \tcp{Decision using row-wise maximum}
        \If {$\max_j(cs_i) > M$} {
          Send intelligent noise vector $\zeta$\;
        } \Else {
          Send $\Delta_{t}$ to server\;
        }
    }
  }
  \caption{Adaptive attack on FoolsGold.\label{alg:adaptive}}
\end{algorithm}
%%%%%%%%%%%%%%%%%%%%%%%%%%%%%%%%%%%%%%%%%%%%%%%%%%%%%%%%%%%%

\begin{figure}[t]
    \centering
    \includegraphics[width=0.8\linewidth]{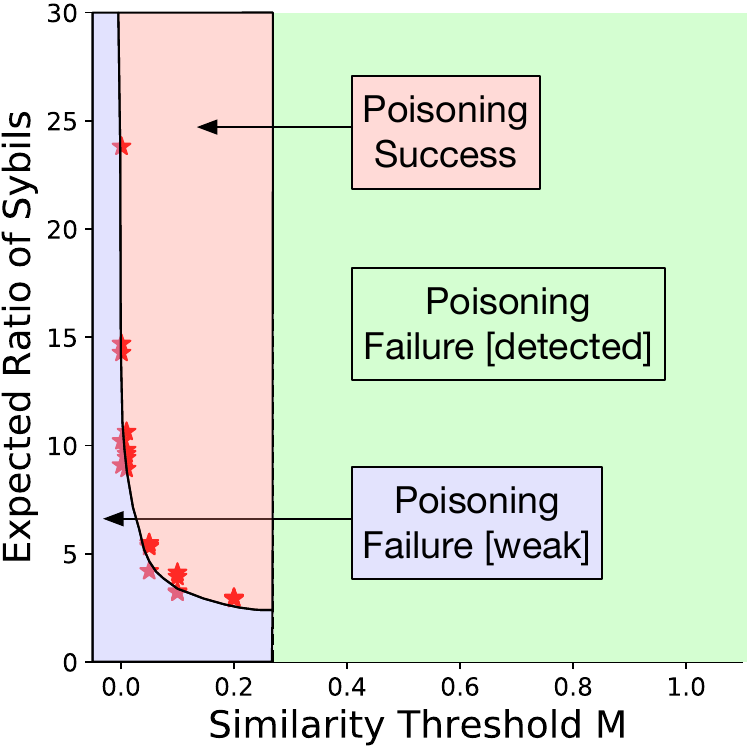}
    \caption{Relationship between similarity threshold and expected
    ratio of sybils per honest opposing client for the adaptive attack
    on FoolsGold (Alg.~\ref{fig:adaptive}) with 2 sybils on MNIST.
    }
    \label{fig:adaptive}
\end{figure}

To demonstrate this, the intelligent noise attack above is
executed by 2 sybils on MNIST, with FoolsGold using the soft weighing
of features in its cosine similarity (the optimal defense for
MNIST against the intelligent noise attack). Figure~\ref{fig:adaptive}
shows the relationship between $M$ and the resulting expected ratio
of sybils needed to match the influence for each honest opposing
client.

For instance, if we observed that the sybils only sent poisoning gradients
25\% of the time, they would need 4 sybils. Given a prescribed
similarity threshold $M$, the values shown are the expected number
of sybils required for the optimal attack. The attack is optimal
because using less sybils does not provide enough influence to poison
the model, while using more sybils is inefficient.

This is shown on Figure~\ref{fig:adaptive} by the three shaded
regions: in the green region to the right ($M > 0.27$), the threshold
is too high and any poisoning attack is detected and removed. In the
blue region on the bottom left, the attack is not detected, but there
is an insufficient number of sybils to overpower the honest opposing
clients. Lastly, in the top left red region, the attack succeeds,
potentially with more sybils than required.

%%%%%%%%%%%%%%%%%%%%%%%%%%%%%%%%%%%%%%%%%%%%%%%%%%%%%%%%%%%%%%%%%%%%%%%%
%%%%%%%%%%%%%%%%%%%%%%%%%%%%%%%%%%%%%%%%%%%%%%%%%%%%%%%%%%%%%%%%%%%%%%%%

With a sufficiently large number of sybils and
appropriately low threshold, attackers can subvert our current defense
for our observed datasets. Finding the appropriate threshold is
challenging as it is dependent on many other factors: the number of
honest clients in the system, the proportion of indicative features
considered by FoolsGold, and the distribution of data.

Furthermore, this attack requires a higher proportion of sybils than
the baseline poisoning attack on federated learning. For example, when
$M$ is set to 0.01, an attacker would require a minimum of 10 sybils
per opposing client to poison the model, whereas in federated
learning, they would only need to exceed the number of opposing
clients. The exact number of sybils required to successfully poison
the model is unknown to attackers without knowledge of the number of
honest clients and their honest training data.

%%%%%%%%%%%%%%%%%%%%%%%%%%%%%%%%%%%%%%%%%%%%%
\subsection{Effects of design elements}
\label{sec:evalmodule}
%%%%%%%%%%%%%%%%%%%%%%%%%%%%%%%%%%%%%%%%%%%%%

Each of the three main design elements (history, pardoning and logit)
described in Section~\ref{sec:design} addresses specific challenges.
In the following experiments we disabled one of the three components and
recorded the training error, attack rate, and target class error of the
resulting model.

\noindent \textbf{History.} The two subversion strategies in
the previous section increase the variance of updates in each
iteration. The increased variance in the updates sent by sybils
cause the cosine similarities at each iteration to be an inaccurate
approximation of a client's sybil likelihood. Our design uses history
to address this issue, and we evaluate it by comparing the
performance of FoolsGold with and without history using an A-5 MNIST
attack with 80\% poisoned data and batch size of 1 (two factors which
were previously shown to have a high variance).

\noindent \textbf{Pardoning.} We claim that honest
client updates may be similar to the updates of sybils, especially if the
honest client owns the data for the targeted class. To evaluate the
necessity and efficacy of our pardoning system, we compare
the performance of FoolsGold on KDDCup with the A-AllOnOne attack with
and without pardoning. 

\noindent \textbf{Logit.} An important motivation for using the
logit function is that adversaries can arbitrarily increase the number
of sybils to mitigate any non-zero weighting of their updates. We
evaluate the performance of FoolsGold with and without the logit
function for the A-99 MNIST attack.

\begin{figure}[t]
    \includegraphics[width=\linewidth]{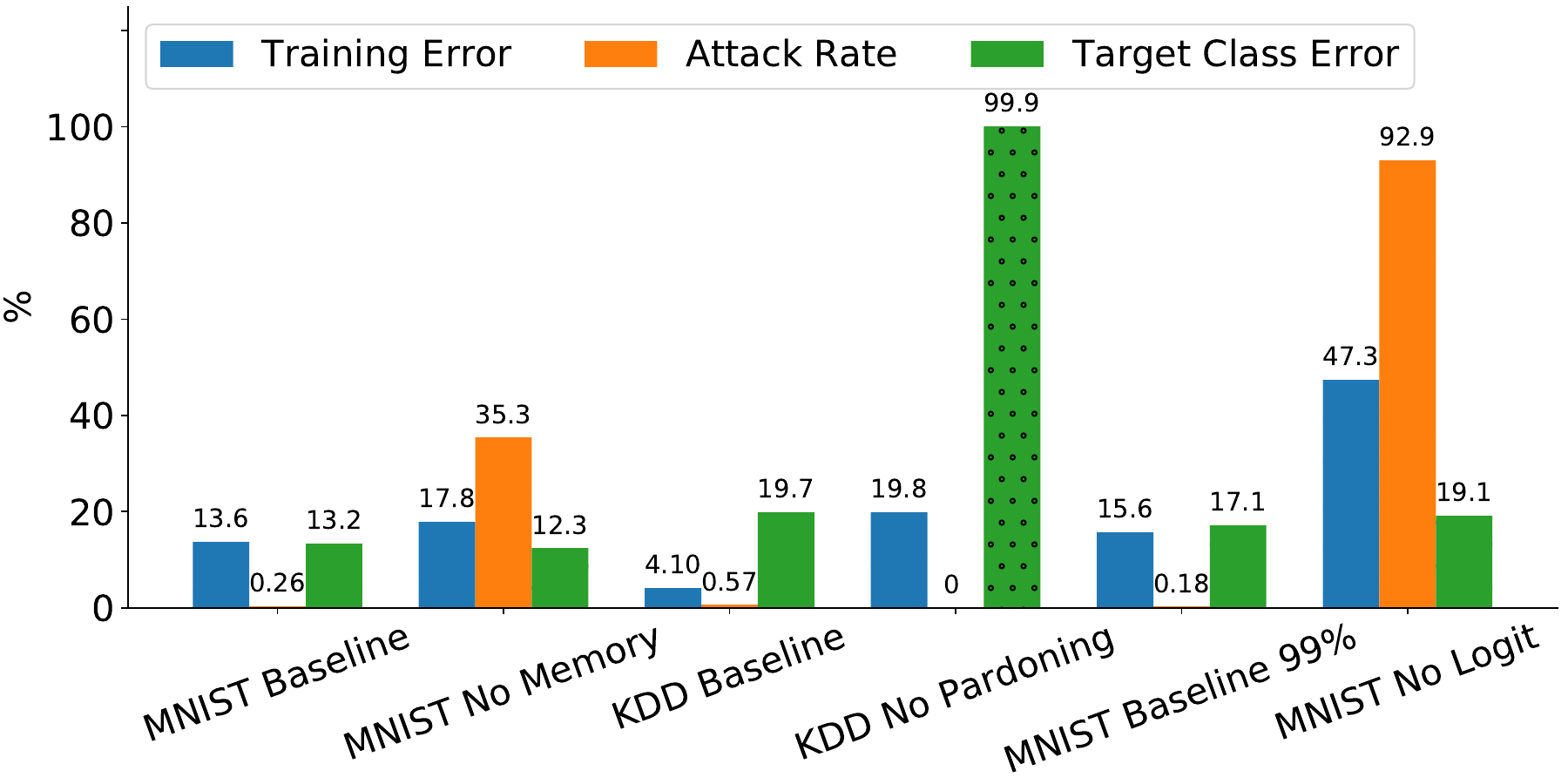}
    \caption{Metrics for FoolsGold with various components
    independently removed.}
    \label{fig:modules}
\end{figure}

Figure~\ref{fig:modules} shows the overall training error, sybil attack
rate, and target class error for the six different evaluations. The
attack rate for the A-AllOnOne KDDCup attack is the average
attack rate for the 5 sets of sybils. 

Overall, the results align with our claims. Comparing the
A-5 MNIST case with and without history, we find that history
successfully mitigates attacks that otherwise would pass through in the
no-history system. Comparing the results of the A-AllOnOne KDDCup
attack, we find that, without pardoning, the training error and target
class error increase while the attack rate was negligible for both
cases, indicating a high rate of false positives for the target class.
Finally, comparing the results for the A-99 MNIST attack, without the
logit function, the adversary was able to mount a successful attack by
overwhelming FoolsGold with sybils, showing that the logit function is
necessary to prevent this attack.

%%%%%%%%%%%%%%%%%%%%%%%%%%%%%%%%%%%%%%%%%%%%%%%%%%%%%%%%%%%%%%%%%%%%%%%%
%%%%%%%%%%%%%%%%%%%%%%%%%%%%%%%%%%%%%%%%%%%%%%%%%%%%%%%%%%%%%%%%%%%%%%%%

%%%%%%%%%%%%%%%%%%%%%%%%%%%%%%%%%%%%%%%%%%%%%
\subsection{FoolsGold performance overhead}
%%%%%%%%%%%%%%%%%%%%%%%%%%%%%%%%%%%%%%%%%%%%%

We evaluate the runtime overhead incurred by augmenting a federated
learning system with FoolsGold. We run the system with and without
FoolsGold with 10 -- 50 clients by training an MNIST classifier on a
commodity CPU and a VGGFace2 deep learning model on a Titan V CPU.

\begin{figure}[t]
    \includegraphics[width=\linewidth]{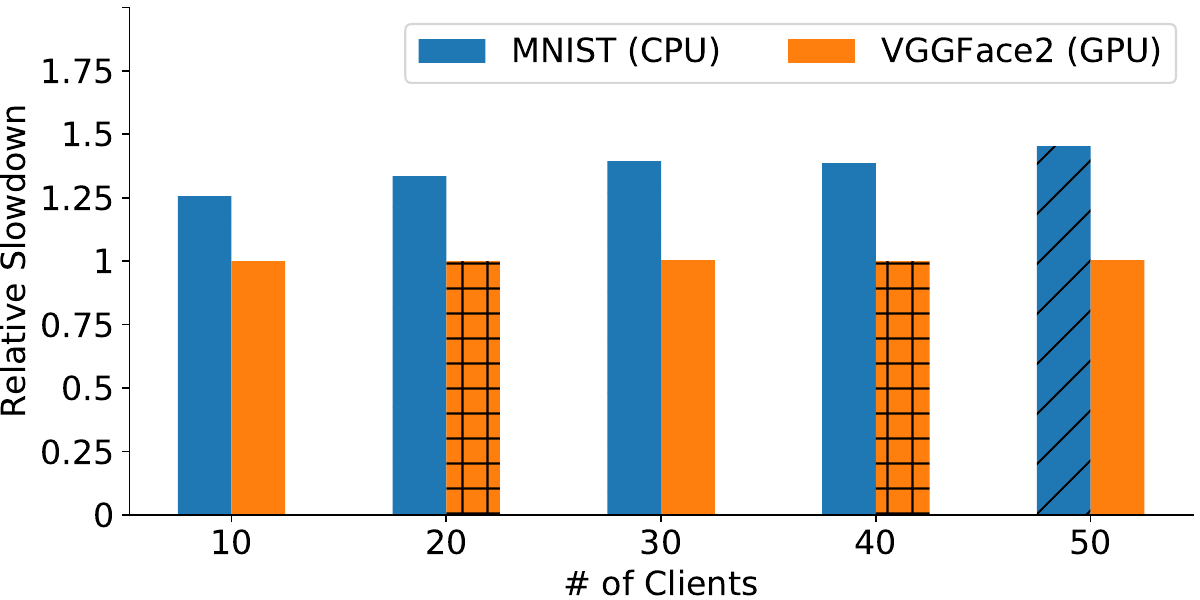}
    \caption{Running time overhead of FoolsGold as compared to
      Federated Learning (baseline of 1), for MNIST (on a CPU) and
      VGGFace2 (on a GPU).}
    \label{fig:runtime}
\end{figure}

Figure~\ref{fig:runtime} plots the relative slowdown added by FoolsGold
for CPU and GPU based workloads. On a CPU, the most expensive part of the
FoolsGold algorithm is computing the pairwise cosine similarity. 
Our Python prototype is not optimized and there are known
optimizations to improve the speed of computing angular distance at
scale~\cite{falconn:2015}. When training a deep learning model on a GPU,
the cost of training is high enough that the relative slowdown from
FoolsGold is negligible. We profiled micro-benchmarks for the total time
taken to execute the FoolsGold algorithm and found that it took less than
1.5 seconds, even with 50 clients.

%%%%%%%%%%%%%%%%%%%%%%%%%%%%%%%%%%%%%%%%%%%%%%%%%%%%%%%%%%%%%%%%%%%%%%%%
%%%%%%%%%%%%%%%%%%%%%%%%%%%%%%%%%%%%%%%%%%%%%%%%%%%%%%%%%%%%%%%%%%%%%%%%

%!TEX root = paper.tex
%%%%%%%%%%%%%%%%%%%%%%%%%%%%%%%%%%%%%%%%%%%%%%%%%%%%%%%%%%%%%%%%%%
\section{Limitations}
\label{sec:limitations}
%%%%%%%%%%%%%%%%%%%%%%%%%%%%%%%%%%%%%%%%%%%%%%%%%%%%%%%%%%%%%%%%%%

\noindent \textbf{Combating a single client adversary.} 
\label{sec:combined}
FoolsGold is designed to counter \emph{sybil-based attacks} and our
results in Figure~\ref{fig:baselines} indicate that FoolsGold is not
successful at mitigating attacks mounted by a single poisoning client.

However, we note that a single malicious actor could be detected
and removed by Multi-Krum~\cite{Blanchard:2017}. Though Multi-Krum
is not designed for and does poorly in the non-IID setting, we
performed an experiment in which FoolsGold was augmented with a
properly parameterized Multi-Krum solution, with $f=1$.

\begin{figure}[t]
    \includegraphics[width=\linewidth]{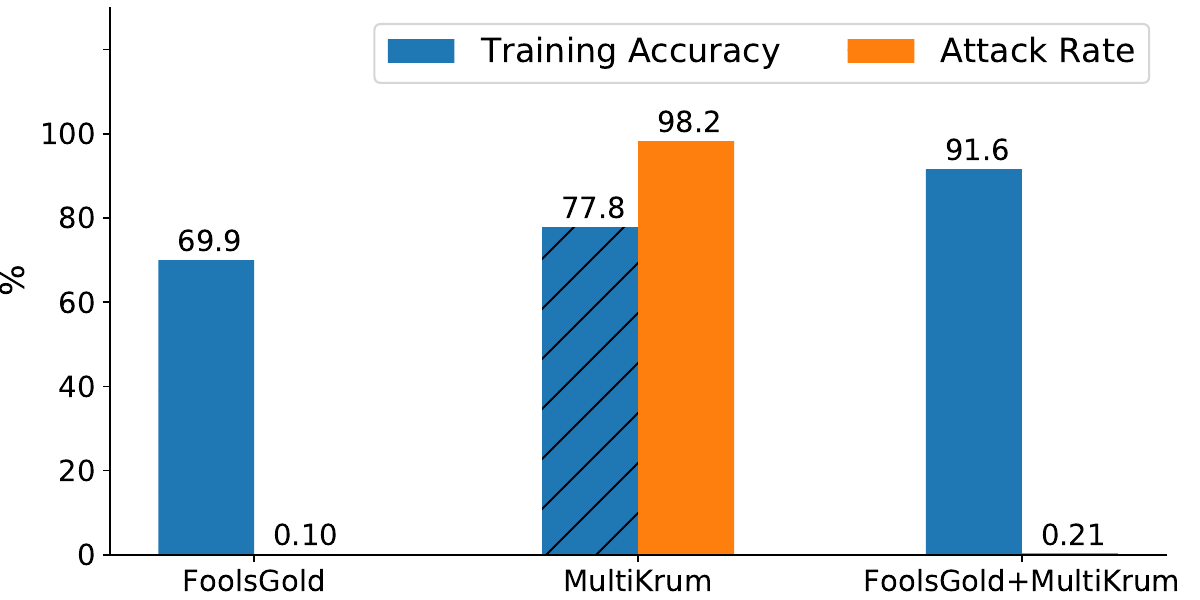}
    \caption{The performance of Multi-Krum with FoolsGold when
    combating the optimal strawman attack on MNIST.} 
    \label{fig:combined}
\end{figure}

We consider an ideal strawman attack in which a single adversarial
client sends the vector to the poisoning objective at each
iteration. This attack does not require sybils and can therefore
bypass FoolsGold. Figure~\ref{fig:combined} shows the training
accuracy and the attack rate for FoolsGold, Multi-Krum, and the two
systems combined when facing a concurrent A-5 and the ideal
strawman attacks.

%%%%%%%%%%%%%%%%%%%%%%%%%%%%%%%%%%%%%%%%%%%%%%%%%%%%%%%%%%%%%%%%%%%%%%%%
%%%%%%%%%%%%%%%%%%%%%%%%%%%%%%%%%%%%%%%%%%%%%%%%%%%%%%%%%%%%%%%%%%%%%%%%

When Multi-Krum uses $f=1$ and FoolsGold is enabled, we see that
Multi-Krum and FoolsGold do not interfere with each other. The
Multi-Krum algorithm prevents the strawman attack, and FoolsGold
prevents the sybil attack. Independently, these two systems fail to
defend both attacks concurrently, either by failing to detect the
ideal strawman attack (against FoolsGold) or by allowing the sybils
to overpower the system (against Multi-Krum).

FoolsGold is specifically designed for handling poisoning attacks from
a group of sybils: we believe current state of the art is better
suited to mitigate attacks from single actors.

\noindent \textbf{Improving FoolsGold against informed attacks.}
In Figure~\ref{fig:adaptive}, we observe that FoolsGold can be
subverted by a knowledgeable adversary with many sybils.
We believe that non-determinism may further help to improve
FoolsGold. For example, by using a weighted random subset of
gradients in the history mechanism or by measuring similarity across
random subsets of client contributions. With more randomness, it
is more difficult for intelligent adversaries to use knowledge about
FoolsGold's design to their advantage.

Another solution is to use a better similarity metric. This includes
strategies from prior work such as incorporating graph-based
similarity~\cite{Yu:2010}, using auxiliary information from
a client dataset~\cite{Viswanath:2015}, or mandating a minimum
similarity score to reject anomalous contributions. While more
informed, these solutions also require more assumptions about
the nature of the attack. 

%!TEX root = paper.tex
%%%%%%%%%%%%%%%%%%%%%%%%%%%%%%%%%%%%%%%%%%%%%%%%%%%%%%%%%%%%%%%%%%
\section{Related work}
\label{sec:related}
%%%%%%%%%%%%%%%%%%%%%%%%%%%%%%%%%%%%%%%%%%%%%%%%%%%%%%%%%%%%%%%%%%

% Additional security work in federated learning has required that
% public key infrastructure exists which
% validates the identity of users ~\cite{Bonawitz:2017}, but prior sybil
% work has shown that relying on public key infrastructure for user
% validation is insufficient~\cite{Viswanath:2015, Wang:2016}. 

We reviewed Multi-Krum~\cite{Blanchard:2017} as the state of the art
defense against poisoning in federated learning. Here we review the broader
literature on secure ML, sybils, and poisoning.

\noindent \textbf{Secure ML.}
Another approach to mitigating sybils in federated learning is to make
the overall system more secure. Trusted execution environments, such
as SGX are an alternative solution~\cite{Ohrimenko:2016}. 
However, we note that sybil-based poisoning can be performed at the
data input level, which is \emph{not} secured by SGX. Even
in a secure enclave running trusted code, a poisoning attack can be
performed through malicious data. FoolsGold can be added to
systems based on SGX to prevent sybil attacks.

Other solutions employ data provenance to prevent poisoning 
attacks~\cite{Baracaldo:2017} by segmenting, identifying, and removing
training data from the model that is likely malicious. This solution
requires extra assumptions about how training data is collected and in
federated learning, where many clients supply data from a variety of
sources, this assumption is unrealistic.

\noindent \textbf{Sybil defenses.}
One way to mitigate sybils is to use proof of
work~\cite{Back:2002}, in which a client must solve a
computationally expensive problem (that is easy to check) to
contribute to the system. Recent alternatives have explored the
use of proof of stake~\cite{Gilad:2017}, which weighs clients by the
value of their \emph{stake} in the system.

Some approach %are not only robust to sybil attacks, but 
actively detect and remove sybils using auxiliary 
information, such as an underlying social network~\cite{Tran:2009}, or
detect and reject malicious
behavior~\cite{Viswanath:2015, Wang:2016}. Many sybil detection strategies
map the interactions between clients into a weighted graph and
use sybil defenses from social networks~\cite{Yu:2010} as a
reduction. However, federated learning limits the amount of
information exposed to the central service, and these defenses may
rely on privacy-compromising information. 
By contrast, FoolsGold does not use any auxiliary information
outside of the learning process.

\noindent \textbf{Other ML poisoning defenses.}
Another poisoning defense involves bagging 
classifiers~\cite{BiggioBag:2011} to mitigate the effects of outliers
in the system. These defenses are effective in scenarios where the
dataset is centralized, but are complex to apply in a
federated learning setting, where access to the data is prohibited.
This algorithm also assumes control of the gradient computation 
during training. % models.

AUROR~\cite{Shen:2016} is a defense designed for the multi-party ML
setting. It defines indicative features to be the most
important model features, and a distribution of modifications to them
is collected and fed to a clustering algorithm. Contributions
from small clusters that exceed a threshold distance are removed. This
clustering assumes that the majority of updates to that
feature come from honest clients for all indicative features.
Unlike AUROR, FoolsGold assumes the presence of sybils and uses gradient
similarity to detect anomalous clients. 

%!TEX root = paper.tex
%%%%%%%%%%%%%%%%%%%%%%%%%%%%%%%%%%%%%%%%%%%%%%%%%%%%%%%%%%%%%%%%%%
\section{Conclusion}
\label{sec:conc}
%%%%%%%%%%%%%%%%%%%%%%%%%%%%%%%%%%%%%%%%%%%%%%%%%%%%%%%%%%%%%%%%%%

The decentralization of ML is driven by
growing privacy and scalability challenges. Federated
learning is a state of the art proposal adopted in
production~\cite{Gboard:2017}. However, such decentralization opens
the door for malicious clients to participate in training.
We considered the problem of poisoning attacks by sybils to achieve a
 poisoning objective. We show that existing defenses to such attacks %to poisoning in federated learning
%% context, such as Multi-Krum,
are ineffective
% when sybils are added
and propose \emph{FoolsGold}, a defense that uses client
\emph{contribution similarity}.

Our results %evaluation results across three different datasets
indicate that FoolsGold can mitigate a variety of attack types and is
effective even when sybils overwhelm honest users. We also considered
advanced attack types in which sybils mix poisoned
data with honest data, add intelligent noise to their updates, and
adaptively rate limit their poisoned updates to avoid detection.
Across all scenarios, our defense is able to outperform prior work.%%  and
%% requires substantially higher number of sybils to defeat honest
%% opposing clients.

FoolsGold minimally changes the federated learning algorithm, relies
on standard techniques such as cosine similarity, and does not require
prior knowledge of the expected number of sybils. We hope that our
work inspires further research on defenses that are co-designed with
the underlying learning procedure.

%% \acks
%% Acknowledgments, if needed.

% The 'abbrvnat' bibliography style is recommended.
\bibliographystyle{IEEEtranS}
\bibliography{paper}

\setcounter{section}{0}
\renewcommand\thesection{Appendix \Alph{section}}
%\renewcommand{\thesection}{Appendix A.}
%!TEX root = paper.tex

%%%%%%%%%%%%%%%%%%%%%%%%%%%%%%%%%%%%%%%%%%%%%%%%%%%%%%%%%%%%%%%%%%%%%%
\section*{Appendix A: Convergence analysis}
\label{sec:converge}
%%%%%%%%%%%%%%%%%%%%%%%%%%%%%%%%%%%%%%%%%%%%%%%%%%%%%%%%%%%%%%%%%%%%%%

\noindent \textbf{Theorem:}
Given the process in Algorithm~\ref{alg:foolsgold}, the convergence
rate of the participants (malicious and honest) is $O(\frac{1}{T^2})$
over $T$ iterations.\\

\noindent \textbf{Proof of theorem:}
We know from the convergence analysis of SGD~\cite{Nemirovski:2009}
that for a constant learning rate, we achieve a $O(\frac{1}{T^2})$
convergence rate.

Let $M$ be the set of malicious clients in the system and $G$ be the
set of honest clients in the system. 
Assume that the adapted learning rates provided at each iteration
$\alpha_i$ are provided by a function $h(i, t)$, where $i$ is the
client index and $t$ is the current training iteration. As long as 
$h(i,t)$ does not modify the local learning rate of the honest
clients and removes the contributions of sybils, the convergence
analysis of SGD applies as if the training was performed with the
honest clients' data. 
\begin{align*}
\forall i &\in M, h(i, t) \rightarrow 0   \label{eq:cond1} 
\tag{cond1} \\
\forall i &\in G, h(i, t) \rightarrow 1 \label{eq:cond2} 
\tag{cond2}
\end{align*}
We will show that, under certain assumptions, FoolsGold satisfies both
conditions of $h(i,t)$. We prove each condition separately.\\

\noindent \textbf{Condition 1:}
Let $v_i$ be the ideal gradient for any given client $i$ from the
initial shared global model $w_0$, that is: $w_0 + v_i = w_i^*$  
where $w_i^*$ is the optimal model relative to any client $i$'s local
training data. Since we have defined all sybils to have the same
poisoning goal, all sybils will have the same ideal gradient, which we
define to be $v_m$.

As the number of iterations in FoolsGold increases, the historical
gradient $H_{i,t}$ for each sybil approaches $v_m$, with error from
the honest client contributions $\epsilon$:
\[
\forall i \in M: \lim_{t \rightarrow \infty} H_{i,t} = v_m +
\epsilon
\]
Since the historical update tends to the same vector for all sybils,
the expected pairwise similarity of these updates will increase as the
learning process continues. As long as the resulting similarity,
including the effect of pardoning between sybils, is below $\beta_m$,
FoolsGold will adapt the learning rate to 0, 
satisfying~\eqref{eq:cond1}.

$\beta_m$ is the point at which the logit function is below
0 and is a function of the confidence parameter $\kappa$:
\[
\beta_m \ge 1 - \frac{e^{-0.5\kappa}}{1 + e^{-0.5\kappa}}
\]
\noindent \textbf{Condition 2:}
Regarding the ideal gradients of honest clients, we assume that 
the maximum pairwise cosine similarity between the ideal
gradient of honest clients is $\beta_g$. As long as $\beta_g$ is
sufficiently low such that FoolsGold assigns a learning rate
adaptation of 1 for all honest clients, the second condition of $h
(i,t)$ is met. $\beta_g$ is the point at which the logit function is
greater than 1 and is also a function of the confidence parameter
$\kappa$:
\[
\beta_g \le 1 - \frac{e^{0.5\kappa}}{1 + e^{0.5\kappa}}
\]
If the above condition holds, FoolsGold will classify these clients to
be honest and will not modify their learning rates. This maintains the
constant learning rate and satisfies~\eqref{eq:cond2}. 

%%%%%%%%%%%%%%%%%%%%%%%%%%%%%%%%%%%%%%%%%%%%%%%%%%%%%%%%%%%%%%%%%%%%%%
\section*{Appendix B: Additional Evaluations}
% \label{sec:extraeval}
%%%%%%%%%%%%%%%%%%%%%%%%%%%%%%%%%%%%%%%%%%%%%%%%%%%%%%%%%%%%%%%%%%%%%%

\noindent\textbf{Comparison to RONI.}
A Reject on Negative Influence (RONI) defense~\cite{Barreno:2010} counters
ML poisoning attacks. When evaluating a set of suspicious training
examples, this defense trains two models: one model using a trusted
dataset, and another model using the union of the trusted and suspicious
data. If the performance of the model degrades the performance beyond a
specified threshold, the data is rejected.

RONI has not been applied to federated learning before. To extend RONI
to a federated learning setting, the server can capture the influence
of a single update, rather than an entire dataset. RONI is ineffective
in this setting because, at any given iteration, an honest gradient
may update the model in an incorrect direction, resulting in a drop in
validation accuracy. This is confounded by the problem that clients
may have data that is not accurately modeled by the RONI validation
set. For example, we can see in Table~\ref{tab:fedattack} that the
prediction accuracy across the other nine digits only dropped slightly
when executing a poisoning attack in a non-IID setting. Depending on
the RONI threshold, this poisoning attack may go undetected.

We show that using RONI with a validation dataset that contains a
uniform distribution of data, is insufficient in countering
sybil-based poisoning in a non-IID setting.

\begin{figure}[t]
    \includegraphics[width=\linewidth]{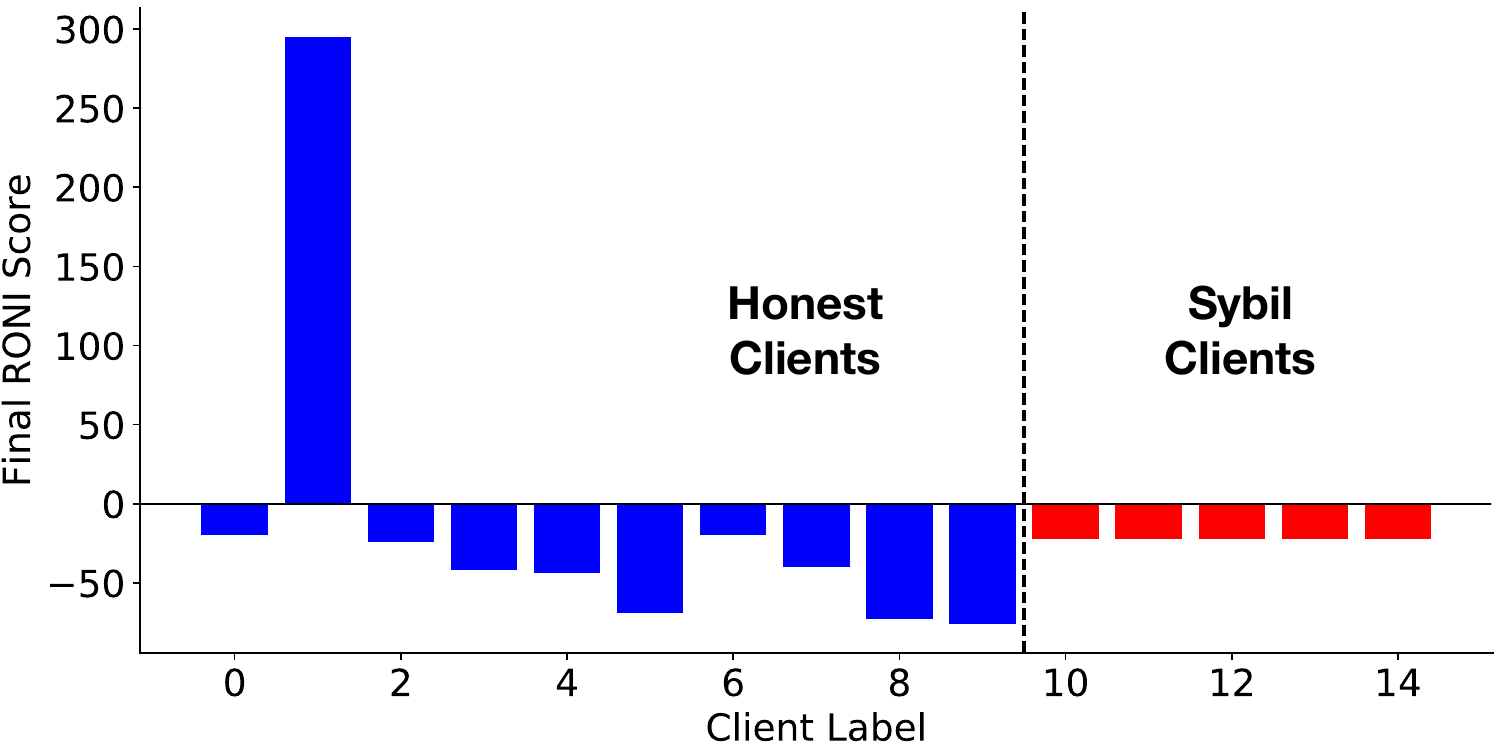}
    \caption{The final sum of the RONI validation scores for the A-5
    attack executed on federated learning: RONI is unable to
    distinguish the malicious clients (5 rightmost) from most of the
    honest ones (the first 10).}
    \label{fig:roni}
\end{figure}

We train an MNIST classifier and use a 10,000 example IID RONI
validation set. We perform an A-5 1-7 attack on the system and
log the total RONI validation score across all 15 clients (10 honest
and 5 sybils) for 3,000 iterations. Figure~\ref{fig:roni} shows the
total sum of the RONI score across all iterations (the 5 right-most
clients in the Figure are sybils). A RONI score below 0 indicates 
rejection. Figure~\ref{fig:roni} shows that all clients
except the honest client with the digit 1 had received a
negative score in every iteration.  

%%%%%%%%%%%%%%%%%%%%%%%%%%%%%%%%%%%%%%%%%%%%%%%%%%%%%%%%%%%%%%%%%%%%%%%%

\begin{figure}[t]
    \includegraphics[width=\linewidth]{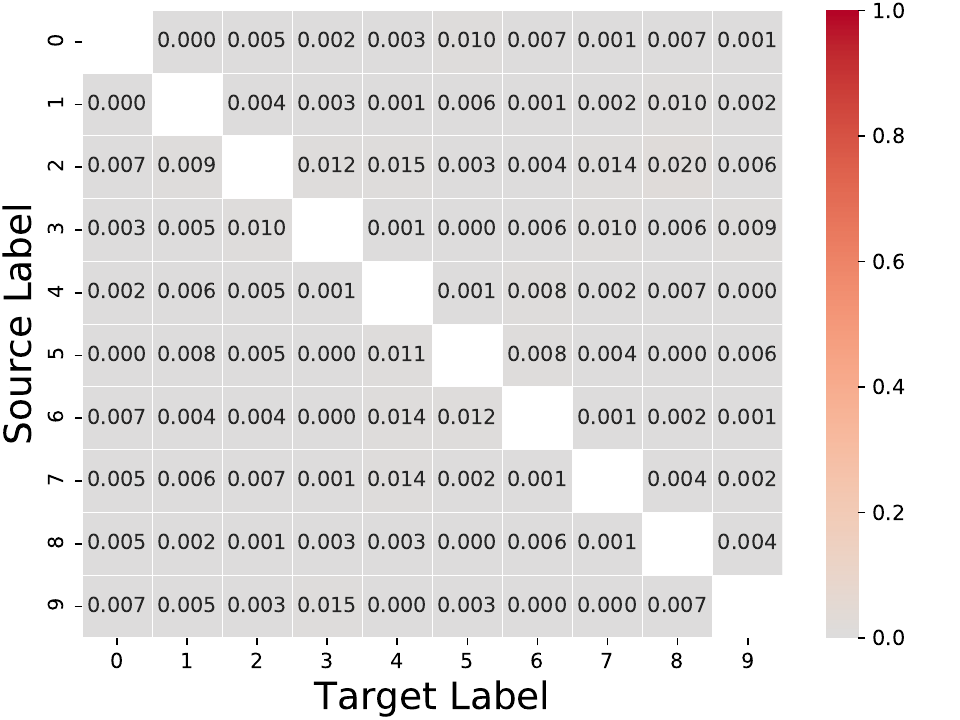}
    \caption{The attack rates for the A-5 attack executed for all
    combinations: attempting to mis-label a true (source) 0-9 class to a
    different (target) 0-9 class.}
    \label{fig:attack_heatmap}
\end{figure}

%%%%%%%%%%%%%%%%%%%%%%%%%%%%%%%%%%%%%%%%%%%%%%%%%%%%%%%%%%%%%%%%%%%%%%%%

In the non-IID setting, clients send updates that do not represent an
update from the global data distribution. Validating individual
updates from a single client in such a fashion produces false
positives because the aggregator holds a validation set that contains
uniform samples, and RONI flags each of their updates as malicious
for doing poorly on the validation set. 

Without prior knowledge of the details of a potential attack, RONI is
unable to distinguish between updates coming from sybils and updates
coming from honest non-IID clients.

\noindent\textbf{Attack generalization.}
In this work, we have used the 1-7 attack on MNIST as our running example.
However, the attacker could be more successful in attacking a
different source and target class. We evaluate FoolsGold's ability to
generalize to other targeted poisoning attacks against MNIST (i.e.,
all other possible source and target MNIST labels) For each possible
source-target combination, we execute the A-5 attack scenario
against FoolsGold.

Figure~\ref{fig:attack_heatmap} shows the resulting attack rate for
our defense across all permutations of the A-5 MNIST poisoning
attacks. 

The highest attack rate observed across all source-target combinations
is 0.02, indicating that FoolsGold generalizes to other class-based
poisoning attacks in MNIST. 

%%%%%%%%%%%%%%%%%%%%%%%%%%%%%%%%%%%%%%%%%%%%%%%%%%%%%%%%%%%%%%%%%%%%%%%%
\begin{figure}[t]
    \includegraphics[width=\linewidth]{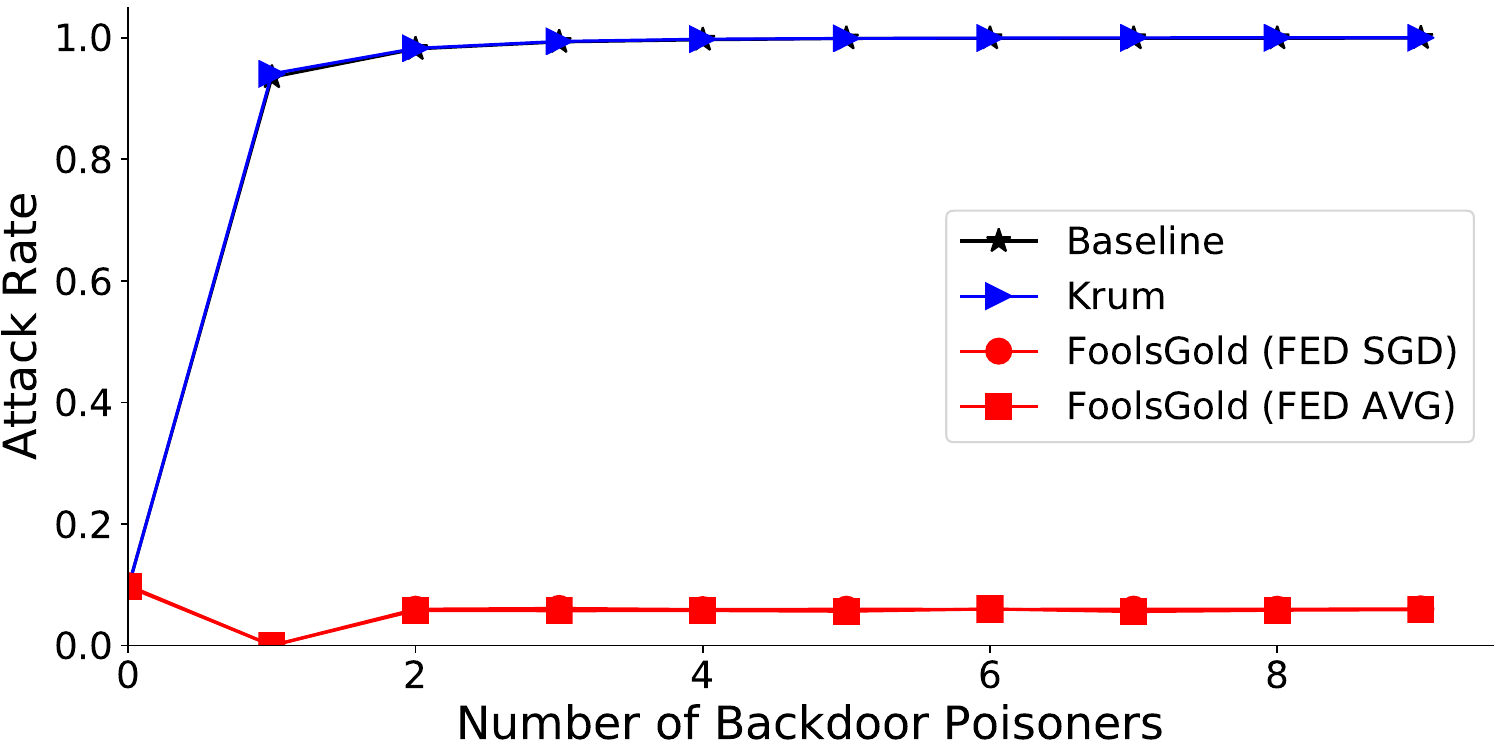}
    \caption{Backdoor attack rate for varying number of sybils, for
      federated learning (Baseline), Multi-Krum, and FoolsGold.}
    \label{fig:backdoor}
\end{figure}
%%%%%%%%%%%%%%%%%%%%%%%%%%%%%%%%%%%%%%%%%%%%%%%%%%%%%%%%%%%%%%%%%%%%%%%%

\noindent\textbf{Backdoor attacks.}
To demonstrate the applicability of FoolsGold towards backdoor sybil
attacks, we repeat the baseline evaluation against the single pixel
backdoor attack~\cite{Gu:2017}. To perform this attack, a random
subset of the MNIST training data is marked with a white pixel in the
bottom-right corner of the image, and labelled as a 7. A successful
backdoor attack results in a model where all images with the backdoor
inserted (a white bottom-right pixel) would be predicted as 7s,
regardless of the other information in the image.

The backdoor attack becomes stronger when a higher proportion of the
training data is poisoned~\cite{Gu:2017}; we view this as identical to
increasing the number of sybils in the system, who all possess a
random subset of the original training data with the backdoor
inserted. The single pixel backdoor was applied with an increasing
number of poisoners (from 0 to 9), against a system with FoolsGold,
Multi-Krum or federated learning. In this experiment, Multi-Krum was
also configured such that the parameter $f$ was equal to the number of
sybils in the system. 

The results of the sybil-based backdoor attack are shown in 
Figure~\ref{fig:backdoor}. FoolsGold is also effective in defending
against the backdoor attack from a large number of sybils, for FEDSGD
and FEDAVG. 

Since the sybils are all performing the same attack, in which all
classes are being backdoored to the same class, the similarities
between these updates will be higher than the similarities between
honest clients.

As in the label-flipping evaluation, there was a high degree of false
positives in FoolsGold when being attacked by only one backdooring
client. While the attack rate was 0, the resulting test accuracy was
below 80\% (due to low prediction rate of 7s).

%%%%%%%%%%%%%%%%%%%%%%%%%%%%%%%%%%%%%%%%%%%%%%%%%%%%%%%%%%%%%%%%%%%%%%%%
\begin{figure}[t]
    \includegraphics[width=\linewidth]{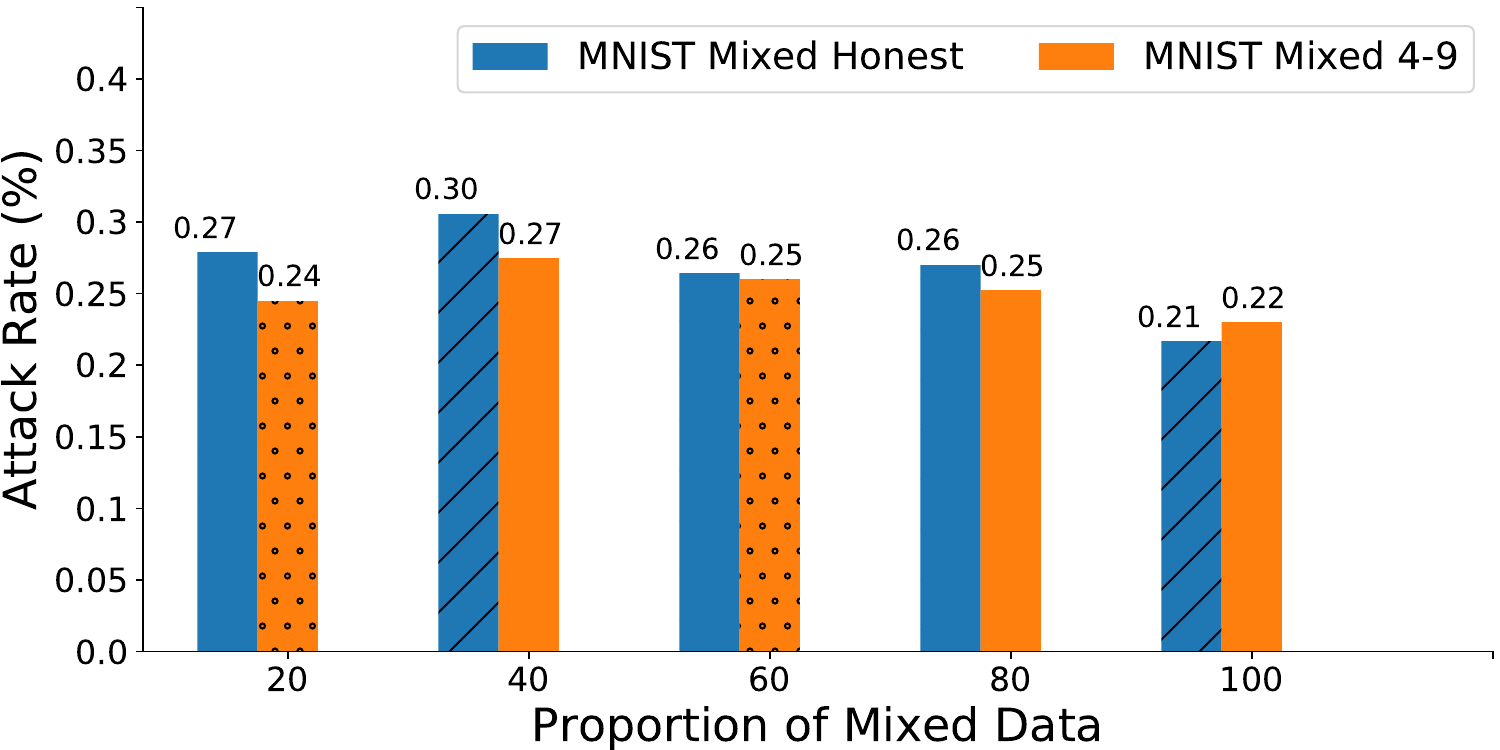}
    \caption{Sybils cannot subvert FoolsGold by
      including a proportion of different honest or attack data.
    }
    \label{fig:subversion}
\end{figure}

\begin{figure}[t]
    \includegraphics[width=\linewidth]{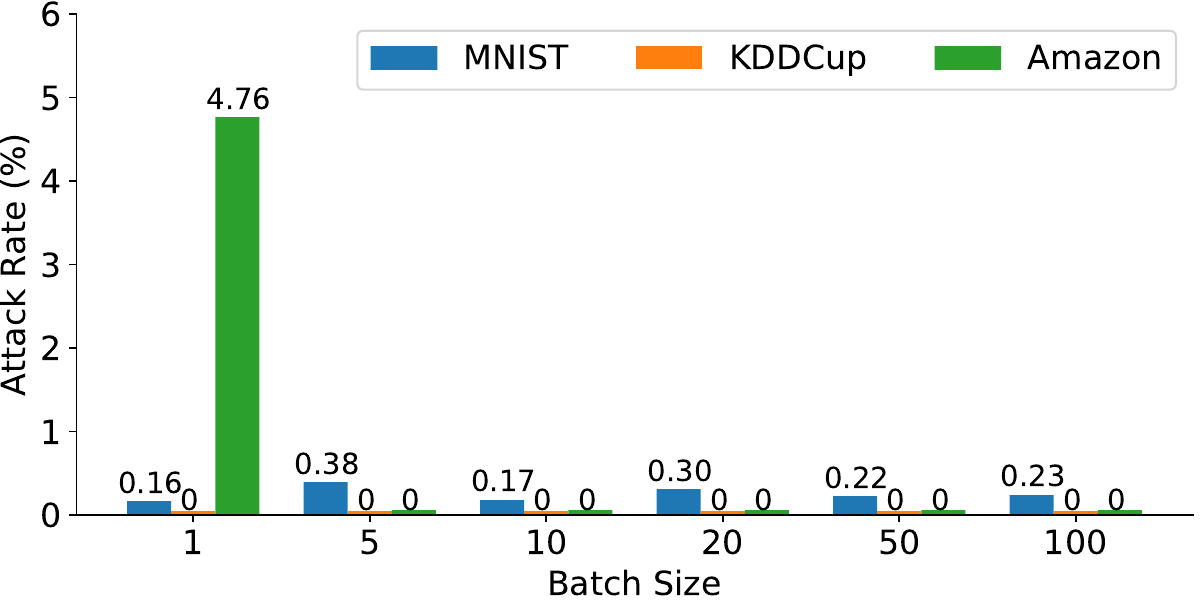}
    \caption{The performance of the A-5 attack on all datasets with
    increasing batch size.}
    \label{fig:batchsize}
\end{figure}
%%%%%%%%%%%%%%%%%%%%%%%%%%%%%%%%%%%%%%%%%%%%%%%%%%%%%%%%%%%%%%%%%%%%%%%%

\noindent \textbf{Resilience to mixed data.}
\balance
Attackers may attempt to subvert our design by creating a dataset with
a mixture of honest and poisoned data. This provides an opportunity
for sybils to appear honest, as they are less similar
when executing SGD. Note that in this case the attacker purposefully
mitigates their attack by labeling a proportion of the poisoned dataset
\emph{correctly}.

To evaluate this attack, we created datasets with 20\%, 40\%, 60\% and
80\% poisoned data for an MNIST 1-7 attack, and the remaining
proportion of the data with honestly labeled data. We also perform a
second experiment in which the attacker mixes their data with poisoned
data for a second 4-9 attack, effectively performing a mixture of two
targeted poisoning datasets simultaneously. Both attacks are mounted
using the \text{A-5} strategy. % against FoolsGold.

Figure~\ref{fig:subversion} plots the attack rate against the
proportion of mixing. This attack was performed with KDDCup
and Amazon datasets, for which all attack rates are 0; we
omit these results. We observe that FoolsGold is robust
to this strategy: the amount of data mixing does not impact
FoolsGold's ability to mitigate sybils, when choosing any proportion
between 20\% and 80\%, the maximum average attack rate observed is
0.003.

At any individual iteration, sybils are more likely to appear as honest
clients. However, as the number of iterations increases, the
average will tend to the same objective, which FoolsGold is able to
detect through its use of history. The effect of this mechanism is
explored more in Section~\ref{sec:evalmodule}. \vspace{.2em}

\noindent\textbf{Resilience to different batch sizes.}
Another factor that impacts FoolsGold's effectiveness is the client
batch size. Given that variance in SGD decreases with more data
points, we expect FoolsGold to perform worse with smaller batch sizes.

To evaluate different batch sizes, we performed A-5 attacks on all
three datasets. In each case, all clients and sybils perform SGD with
the same batch size for values of 1, 5, 10, 20, 50, and 100. Since no
Amazon data partition contains over 50 examples, we do not evaluate
Amazon with batch sizes of 50 and 100.

Figure~\ref{fig:batchsize} shows the attack rate as the batch size
increases. It shows that FoolsGold is resilient to the batch size for
A-5 attacks performed on MNIST and KDDCup, achieving an attack
rate at or near 0.

The only instance in which performance of the system suffered was for
the A-5 Amazon attack with the batch size set to 1; the resulting
attack rate was 4.76\%. This is due to the curse of dimensionality:
there is a higher variance in the similarities between updates when
the dimension size is large (10,000). This variance is maximal when the
batch size is lowest.

% The bibliography should be embedded for final submission.
%% \begin{thebibliography}{}
%% \softraggedright

%% \bibitem[Smith et~al.(2009)Smith, Jones]{smith02}
%% P. Q. Smith, and X. Y. Jones. ...reference text...
%\end{thebibliography}

\end{document}